\documentclass[lettersize,journal,hidelinks]{IEEEtran}
\usepackage[T1]{fontenc}
\usepackage{amsmath,amsfonts,amssymb}
\usepackage{dsfont}
\usepackage{array}
\usepackage{textcomp}
\usepackage{stfloats}
\usepackage{placeins}
\usepackage{url}
\usepackage{verbatim}
\usepackage{graphicx}
\usepackage{todonotes}
\usepackage{orcidlink}
\usepackage[style=ieee, doi=true, url=true, isbn=false, minnames=1, maxnames=10, date=year, maxcitenames=2, hyperref=true]{biblatex}
\usepackage{makecell}
\usepackage{booktabs}
\usepackage{multirow}
\usepackage{lettrine}
\usepackage{csquotes}
\usepackage{mathtools}
\usepackage{xfrac}
\usepackage{lipsum}
\usepackage{algorithm}
\usepackage{algpseudocode}
\usepackage{diagbox}
\usepackage{caption}
\usepackage{subcaption}
\usepackage{adjustbox}
\usepackage{tikz}
\usepackage[capitalise,noabbrev]{cleveref} 
\usepackage{xcolor, colortbl}
\usepackage{siunitx}

\newcommand{\tbf}[1]{\textbf{#1}}
\newcommand{\tul}[1]{#1}
\newcommand{\newpagetoggle}{}

\Crefname{figure}{Fig.}{Figs.}
\usepackage{pifont}

\newcommand{\cmark}{\ding{51}}
\newcommand{\xmark}{\ding{55}}

\newcommand{\aup}{$\uparrow$}
\newcommand{\adown}{$\downarrow$}

\usepackage{xspace}
\newcommand{\soso}{\textit{S1}$\to$\textit{S1}\xspace}
\newcommand{\stst}{\textit{S2}$\to$\textit{S2}\xspace}
\newcommand{\sost}{\textit{S1}$\to$\textit{S2}\xspace}
\newcommand{\stso}{\textit{S2}$\to$\textit{S1}\xspace}

\usepackage[commandnameprefix=always,final]{changes}
\usepackage{acronym}
    \acrodef{AA}{average accuracy}
    \acrodef{AI}{artificial intelligence}
    \acrodef{CBIR}{content-based image retrieval}
    \acrodef{CL}{contrastive learning}
    \acrodef{CSMAE}{Cross-Sensor Masked Autoencoder}
    \acrodef{CSMoE}{Cross-Sensor Mixture-of-Experts}
    \acrodef{CV}{computer vision}
    \acrodef{CZ}{climate zone}
    \acrodef{DL}{deep learning}
    \acrodef{EO}{earth observation}
    \acrodef{FLOP}{floating-point operation}
    \acrodef{FFN}{feed-forward network}
    \acrodef{FM}{foundation model}
    \acrodef{GAP}{global average pooling}
    \acrodef{GSD}{ground sample distance}
    \acrodef{IoU}[IoU]{mean intersection-over-union}
    \acrodef{KNN}[k-NN]{k-nearest neighbors}
    \acrodef{LULC}{land-use/land-cover}
    \acrodef{MAE}{masked autoencoder}
    \acrodef{mAP}[mAP$_\mu$]{micro-mean average precision}
    \acrodef{MiM}{mutual information maximization}
    \acrodef{MIM}{masked image modeling}
    \acrodef{MLP}{multi-layer perceptron}
    \acrodef{MoE}{mixture-of-experts}
    \acrodef{MTC}[MTom]{Major TOM Core}
    \acrodef{mTC}[MTom$_\mu$]{Minor TOM Core}
    \acrodef{rTC}[MTom$_r$]{Minor TOM Core (random)}
    \acrodef{NTXent}[NT-Xent]{normalized temperature-scaled cross entropy}
    \acrodef{PIMask}{patch-incomplete masking}
    \acrodef{RS}{remote sensing}
    \acrodef{S1}{Sentinel-1}
    \acrodef{S2}{Sentinel-2}
    \acrodef{SAR}{synthetic aperture radar}
    \acrodef{SOTA}{state-of-the-art}
    \acrodef{SSL}{self-supervised learning}
    \acrodef{UMR}{intra-modal reconstruction loss}
    \acrodef{CMR}{cross-modal reconstruction loss}
    \acrodef{ViT}{vision transformer}
    \acrodef{VQA}{visual question answering}
    
    \acrodefplural{MoE}[MoEs]{mixture-of-experts}

\IEEEoverridecommandlockouts

\usepackage{etoolbox}
\newtoggle{highlightchanges}
\togglefalse{highlightchanges} 

\AtBeginEnvironment{tabular}{\iftoggle{highlightchanges}{\color{blue}}{}}
\AtBeginEnvironment{algorithmic}{\iftoggle{highlightchanges}{\color{blue}}{}}
\newcommand{\applyhighlighting}{%
  \iftoggle{highlightchanges}{%
    \color{blue}%
    \captionsetup{font={color=blue}}%
  }{}%
}

\begin{document}
\sisetup{range-phrase=\,--\,}  
\newcommand{\titleName}{CSMoE: An Efficient Remote Sensing Foundation~Model with Soft~Mixture-of-Experts}
\title{\titleName}

\author{
Leonard Hackel\orcidlink{0000-0002-5831-1237},~\IEEEmembership{Graduate~Student~Member,~IEEE},
Tom Burgert\orcidlink{0000-0003-3453-2729},~\IEEEmembership{Member,~IEEE},
and~Begüm~Demir\orcidlink{0000-0003-2175-7072},~\IEEEmembership{Senior~Member,~IEEE}
    \thanks{
        The authors
        are with the Berlin Institute for the Foundations of Learning and Data (BIFOLD) and
        Technische Universit\"at Berlin, 10623 Berlin, Germany (emails: l.hackel@tu-berlin.de (corresponding author), t.burgert@tu-berlin.de, demir@tu-berlin.de).
    }
}

\markboth{submitted to arXiv}%
{L. Hackel, T. Burgert and B. Demir: \titleName}

\IEEEpubid{0000--0000/00\$00.00~\copyright~2026 arXiv}

\maketitle

\applyhighlighting

\begin{abstract}
\Ac{SSL} through \acp{MAE} has recently attracted great attention for \ac{RS} \ac{FM} development, enabling improved representation learning across diverse sensors and downstream tasks.
However, existing \ac{RS} \acp{FM} often either suffer from substantial computational complexity during both training and inference or exhibit limited representational capacity.
\chadded{In addition, their pretraining datasets can contain redundant images, which increases training cost without improving the learned representations.
These issues restrict their practical applicability in \ac{RS}.
To address these limitations, we improve the computational efficiency of \ac{RS} \acp{FM} along two axes: i)~model efficiency; and ii)~training-data efficiency.
Model efficiency is achieved by integrating the Soft \ac{MoE} mechanism into the \ac{FM}, which allows modality-specific expert processing alongside shared cross-sensor representation learning while reducing computational complexity at both training and inference time.
We apply this adaptation to the \ac{CSMAE} model, which serves as our main baseline, resulting in the \ac{CSMoE} model.
Training-data efficiency is achieved by a thematic-climatic descriptor-driven sampling strategy, which constructs a reduced training set from a large-scale image archive while retaining its geographic and thematic-climatic diversity, and thus reduces pretraining cost.}
Extensive experiments on scene classification, semantic segmentation, and \ac{CBIR} show that \ac{CSMoE} remains competitive with state-of-the-art \ac{RS} \acp{FM} while requiring substantially fewer \acp{FLOP}.
The associated code for the model and the training set creation, as well as the pretrained model weights, will be available at \url{https://git.tu-berlin.de/rsim/csmoe}.
\end{abstract}

\acresetall
\section{Introduction}\label{sec:intro}
\acresetall

\lettrine{W}{ith} the advances in \ac{SSL} and the increasing availability of large-scale \ac{EO} data,
the development of \acp{FM} has attracted great attention in the \ac{RS} community for representation learning problems~\cite{szwarcman2024prithvi,bastani2023satlaspretrain,wang2023ssl4eo,sun2022ringmo,cong2022satmae,reed2023scale,xiong2024dofa,jakubik2025terramind,hackstein2024exploring}. 
\acp{FM} aim to learn general-purpose, task-agnostic representations that can process data from diverse sensors and solve downstream tasks with minimal fine-tuning. 
Unlike conventional \ac{DL} models in \ac{RS} that are often tailored to specific tasks and data modalities, \ac{RS} \acp{FM} aim for broader generalization. 
This is achieved by relying on \ac{SSL}-based learning objectives such as reconstruction-based (e.g., \ac{MIM}~\cite{He2022mae}) or contrastive (e.g., MOCO~\cite{he2020moco} and DINOv2~\cite{oquab2023dinov2}) learning for large-scale pretraining using a large amount of unlabeled data.
The design and development of \acp{FM} has mainly evolved along three primary axes:
i) scaling up model size and training data to increase representational capacity~\cite{szwarcman2024prithvi, bastani2023satlaspretrain, francis2024major}; 
ii) introducing architectural innovations to accurately represent the complex content of \ac{RS} images~\cite{sun2022ringmo, cong2022satmae,reed2023scale}; 
and iii) integrating multiple \ac{EO} modalities to enhance multi-modal and cross-sensor characteristics~\cite{xiong2024dofa, jakubik2025terramind}. 

There are several \acp{FM} developed in \ac{RS}, focusing on scaling model and dataset sizes to improve generalization of learned representations. 
As an example, Prithvi~V2~\cite{szwarcman2024prithvi} scales to more than 600 million parameters and is trained on global time-series data from \ac{S2} and Landsat-8/9 satellites, enabling improved performance on tasks such as disaster response and ecosystem monitoring. 
Satlas~\cite{bastani2023satlaspretrain} introduces a multi-task dataset with over 300 million annotations. 
SSL4EO~\cite{wang2023ssl4eo} complements these efforts with a global, seasonally diverse pretraining dataset, while Major TOM~\cite{francis2024major} contributes an extensible framework based on a geographical indexing system, and introduces a multi-modal dataset (called \ac{MTC}) with up to 23 TB per image modality.
Beyond scale, architectural innovations have emerged to better model the complex content of \ac{RS} images.
As an example, RingMo~\cite{sun2022ringmo} introduces a generative masking strategy for \acp{MAE} tailored to extracting fine-grained features from \ac{RS} images.
As another example, SatMAE~\cite{cong2022satmae} incorporates temporal and spectral encodings, while ScaleMAE~\cite{reed2023scale} leverages resolution-aware positional embeddings for improved cross-scale performance. 
Progress has also been made in developing modality-agnostic \acp{FM}.
For example, DOFA~\cite{xiong2024dofa} employs a wavelength-conditioned dynamic patch embedding layer to accommodate different channel configurations, thereby supporting the change of the image modality at the time of inference without retraining.
TerraMind~\cite{jakubik2025terramind}, on the other hand, enables cross-modal generation through a dual-scale architecture and modality-conditioned decoding. 
All these developments reflect a growing trend toward scalable, general, and multi-modal \acp{FM} in \ac{RS}.
For a comprehensive overview of \acp{FM} in \ac{RS}, we refer the reader to \cite{jiao2023brain, xiao2025foundation, lu2025fmsurvey}. 
\IEEEpubidadjcol

However, this progress comes at a significant computational cost. 
While \acp{FM} benefit from a high representational capacity, reflected in a large number of parameters that enable them to model complex functions, this capacity is often accompanied by substantial computational complexity, commonly measured in \acp{FLOP}, needed for inference of an image.
The importance of computational efficiency has already been recognized in various \ac{RS} tasks, such as scene classification~\cite{chen2022rscnet,cheng2020efficientclassification,xu2021efficientSeg}, semantic segmentation~\cite{xu2021efficientSeg,WANG2022unetformer}, and \ac{VQA}~\cite{hackel2023lit4rsvqa,hackel2024configilm}. 
By contrast, such efficiency-oriented approaches remain largely unexplored in the context of \acp{FM}.
\Cref{tab:model_flop_comparison} shows the computational complexity and representational capacity of different \acp{FM} as well as their ability to process multiple data modalities and the size of their pretraining archive.
By analyzing the table, one can see that some \acp{FM} focus on maximizing representational capacity (e.g., Prithvi~V2-600 with over 630 million parameters).
However, their computational complexity during inference exceeds 160 billion \acp{FLOP} per image. 
Others, such as Satlas, operate at a lower computational budget but offer limited representational capacity with only 88 million parameters. 
Models like DOFA and TerraMind introduce multi-modal learning capabilities, yet their per-sample inference cost remains high (approximately 17.5 billion \acp{FLOP}), indicating that improvements in generalization often come with significant computational resource demands. 
\chadded{
This computational cost is particularly critical for \ac{RS} tasks in which an \ac{FM} is applied to each image of a large-scale archive.
As an example, in \ac{CBIR} the feature vector of every archive image is extracted before any query can be answered, and thus the inference cost of an \ac{FM} scales with the size of the archive.
Note that the computational cost of \ac{RS} \acp{FM} applies not only at inference time, but also during pretraining, where it is largely driven by the size and the redundancy of the pretraining archive.}
\begin{table}[t]
\centering
\setlength\tabcolsep{3pt}
\caption{Comparison of \ac{RS} \acp{FM} regarding their model sizes as number of parameters (\#P), computational complexity in \acp{FLOP}, and number of pixels in the pretraining archive (PT Pix.). The \ac{FLOP}-calculation is based on a forward pass of a single S2-image (at 224$\times$224 pixels with the subset of bands supported by the respective model) for feature extraction. 
Evaluation based on reference implementations in TerraTorch~\cite{gomes2025terratorch}. M = Million/Mega, G = Billion/Giga, T = Trillion/Tera.}
\label{tab:model_flop_comparison}
\begin{tabular}{lccccc}
\toprule
\textbf{Model} & \textbf{\makecell{multi-\\modal}} & \textbf{\#P} \aup & \textbf{FLOPs} \adown & \textbf{\makecell{PT\\Pix.} \adown} \\
\midrule
Prithvi V2-300 \cite{szwarcman2024prithvi} & \xmark                             & \tul{304M} &     59.85G  &      210.7G  \\  
Prithvi V2-600 \cite{szwarcman2024prithvi} & \xmark                             &      631M  &    162.18G  &      210.7G  \\  
Satlas \cite{bastani2023satlaspretrain}    & \xmark\footref{fn:multimodal_text} &       88M  &     17.12G  &       14.6T  \\  
DOFA \cite{xiong2024dofa}                  & \cmark                             &      111M  &     17.47G  &       20.6G  \\  
TerraMind \cite{jakubik2025terramind}      & \cmark                             &       87M  &     17.84G  &      451.6G  \\  
CSMAE \cite{hackstein2024exploring}        & \cmark                             &       87M  &      5.64G  &        3.9G  \\  
CSMoE (ours)                               & \cmark                             &      271--277M  & 2.92 -- 13.40G &  \tul{14.5G} \\
\bottomrule
\end{tabular}
\end{table}

\footnotetext[1]{\label{fn:multimodal_text}Different versions of the model exist for individual modalities, but no unified model for multiple modalities.}

\addtocounter{footnote}{1}

\chadded{
The pretraining archives of recent \ac{RS} \acp{FM} comprise hundreds of billions of pixels~\cite{dias2024oreolefm}, and thus their construction and processing require substantial computational resources.
These archives can contain redundant images, such as images from the same \ac{LULC} class under the same \ac{CZ} (e.g., extensive deserts or forests)~\cite{szwarcman2024prithvi, wang2023ssl4eo, roscher2024better, stanimirova2023global}.
When included in the training set, such images may not contribute significant new information~\cite{Kerdreux_2025_CVPR}.
Accordingly, a high amount of redundancy increases the pretraining cost without improving the capability of the learned representations.
Recent studies address this issue in general machine learning through automatic data selection and curation strategies (e.g., graph-based~\cite{van2024graph} and clustering-based~\cite{vo2024automatic} approaches), while such strategies remain underexplored for \ac{RS} \acp{FM}.
The above-mentioned issues result in the computational inefficiency of current \ac{RS} \acp{FM}, which applies during pretraining as well as at inference time.
}

\chadded{To address these issues, we improve computational efficiency along two axes: i)~model efficiency; and ii)~training-data efficiency. 
The first axis is achieved by injecting a \ac{MoE} into an \ac{RS} \ac{FM} to reduce computational complexity at both training and inference time. 
The second axis is achieved by a sampling strategy that reduces training cost by decreasing the size and redundancy of the pretraining archive.}
Although prior studies have explored the use of \acp{MoE} in \ac{RS} model development~\cite{hui2025rsmoe, bi2025ringmoe,liu2024rsunivlm,chen2025generalizable}, these works primarily employ \acp{MoE} as a means to scale model capacity, rather than to explicitly target computational efficiency.
In contrast, our focus is on leveraging \acp{MoE} to enhance efficiency, for which we adopt the Soft \ac{MoE}~\cite{puigcerver2024softmoe}, a variant designed to combine high representational capacity with reduced computational cost.
When Soft \acp{MoE} are applied in transformers~\cite{vaswani2017attention, dosovitskiy2021an}, each input token is softly routed to multiple expert branches, and their outputs are combined using the learned routing weights.
The mechanism is particularly well-suited for efficient \ac{RS} \acp{FM}, as it combines a lightweight routing mechanism with soft token mixing, thereby increasing representational capacity while reducing computational complexity.
We inject the Soft \ac{MoE} into the \ac{CSMAE}~\cite{hackstein2024exploring} \ac{FM} to create our \ac{FM} that we call \ac{CSMoE}.
\chadded{Accordingly, we consider \ac{CSMAE} as our main baseline throughout this work.
For the second axis, we introduce the thematic-climatic descriptor-driven sampling strategy, which selects a representative training set from a large-scale image archive while retaining the geographic and thematic-climatic diversity of the original archive.}
To this end, our proposed thematic-climatic descriptor-driven sampling strategy consists of two stages:
The first stage aims at preserving thematic-climatic diversity while reducing the sample size.
This is achieved by assigning each sample a combination of climatic and thematic descriptors (e.g., a \ac{CZ} and a set of \ac{LULC} classes).
Then, a fixed number of samples is drawn from each combination of thematic-climatic descriptors to build an initial subsampled training set.
In the second stage, the aim is to enhance the spatial diversity of the subset.
This is achieved by applying a genetic algorithm~\cite{holland1992genetic} that maximizes the geodesic distance within the selected samples.
Through these two stages, the strategy reduces redundancy by discarding semantically similar samples (e.g., multiple samples from the same combination of climate and thematic product descriptors) while retaining the geographic and thematic-climatic diversity of the original archive.
This reduces the training time and environmental impact of \ac{FM} training in \ac{RS}.
We train our \ac{CSMoE} model using a subset of \ac{MTC}~\cite{francis2024major}, where we select a training set using our thematic-climatic descriptor-driven sampling strategy.
Through extensive experiments, we show that our \ac{CSMoE} model achieves a downstream performance competitive to that of state-of-the-art \acp{FM}, while significantly reducing inference cost.
\chadded{As an example, our \ac{CSMoE} model reaches a \ac{mAP} comparable to that of Prithvi~V2-600 on m-bigearthnet while requiring an order of magnitude fewer \acp{FLOP}.}
Our results underscore \ac{CSMoE}'s efficiency and strong generalization capability across diverse \ac{RS} tasks.

The main contributions of this work are summarized as follows:
\begin{itemize}
  \item According to our knowledge, \ac{CSMoE}, which injects a Soft \ac{MoE} into \ac{CSMAE}, is the first computationally efficient \ac{FM} in \ac{RS}.
  As shown in \cref{tab:model_flop_comparison}, this integration increases the model's representational capacity while significantly reducing computational complexity during both training and inference compared to models of similar size.
  \item \chadded{To efficiently pretrain the \ac{CSMoE} model, we introduce a thematic-climatic descriptor-driven sampling strategy that constructs a reduced training set from a large-scale unlabeled \ac{RS} archive, while retaining its geographic and thematic-climatic diversity.}
  \item We conduct extensive experiments on a suite of single- and multi-label scene classification and semantic segmentation (pixelwise classification) benchmarks, as well as both uni-modal and cross-modal \ac{CBIR} tasks in \ac{RS} and compare our \ac{CSMoE} model with state-of-the-art \acp{FM} in \ac{RS}.
  We demonstrate that \ac{CSMoE} has a comparable or superior downstream performance while requiring fewer computational resources than competing models.
\end{itemize}

The remainder of this paper is organized as follows:
In \cref{sec:method}, we introduce the proposed \ac{CSMoE} model and our efficient thematic-climatic descriptor-driven sampling strategy.
\Cref{sec:experiment} describes the experimental setup, including the construction of the pretraining dataset via the proposed sampling strategy, as well as the downstream datasets and implementation details.
\Cref{sec:evaluation} presents a comprehensive evaluation of the \ac{CSMoE} model on scene classification, semantic segmentation, and image retrieval tasks, along with a sensitivity analysis.
Finally, \cref{sec:conclusion} concludes the paper.

\newpagetoggle


\section{Proposed Efficient Remote Sensing Foundation~Model with Soft~Mixture-of-Experts}\label{sec:method}
Let $\mathcal{X}$ and $\mathcal{Y}$ denote two co-registered multi-modal \ac{RS} image archives associated with different image modalities (i.e., acquired by different sensors). 
The archives $\mathcal{X} = \{\boldsymbol{x}_i\}_{i=1}^N$ and $\mathcal{Y} = \{\boldsymbol{y}_i\}_{i=1}^N$ each include $N$ images, where $\boldsymbol{x}_i$ and $\boldsymbol{y}_i$ are the $i$th \ac{RS} images in the respective archives and $(\boldsymbol{x}_i, \boldsymbol{y}_i)$ is the $i$th multi-modal image pair that includes two images acquired by different sensors on the same geographical area.
We assume that an unlabeled training set $\mathcal{T}=\{(\boldsymbol{x}_i,\boldsymbol{y}_i)\}_{i=1}^N$ is available for representation learning.
Although our approach can be extended to multiple modalities by applying the proposed processing steps in a pairwise combination, we focus on the dual-modality case to simplify the mathematical treatment in this paper.

The computational complexity of current \ac{RS} \acp{FM} during both training and inference poses significant challenges for operational deployment.
To address this limitation, we propose to integrate the \ac{MoE} mechanism into existing \ac{MAE}-based \ac{FM} architectures to achieve improved computational efficiency while maintaining or increasing representational capacity.
For representation learning from the training set $\mathcal{T}$, \ac{MIM} (i.e., \ac{SSL} through \acp{MAE}) has recently emerged as one of the most successful approaches in \ac{RS}. 
Among various \ac{MAE} variants, the \ac{CSMAE} model~\cite{hackstein2024exploring} has shown promising results in both uni-modal and cross-modal representation learning.
However, like all existing multi-modal \ac{MAE}-based \acp{FM}, \ac{CSMAE} suffers from a high computational complexity during training and inference with limited representational capacity.
As a first time in \ac{RS}, we explore the effectiveness of integrating the Soft \ac{MoE} mechanism into \acp{MAE} to create computationally efficient \acp{FM}.
To this end, we select \ac{CSMAE} as our base model due to its proven cross-modal capabilities and apply our \ac{MoE} injection adaptation to create \ac{CSMoE}. 
In the following subsections, we first provide background information on the \ac{CSMAE} model, and then present our adaptation that injects the Soft \ac{MoE} into \ac{CSMAE} and our efficient thematic-climatic descriptor-driven sampling strategy.

\subsection{Basics on Cross-Sensor Masked Autoencoder}\label{sec:csmae}
Given an image $\boldsymbol{x} \in \mathcal{X}$ or $\boldsymbol{y} \in \mathcal{Y}$, \acp{MAE} operate by dividing the image into $P$ non-overlapping image patches of size $\rho \times \rho$, forming a token set $\mathcal{P} = \{\boldsymbol{p}_n\}_{n=1}^P$. 
A random subset $\mathcal{M} \subset \{1, \ldots, P\}$ is masked, while the remaining tokens $\mathcal{U} = \{1, \ldots, P\} \setminus \mathcal{M}$ are passed to a transformer encoder $e$ to produce latent representations $\mathcal{Z}_{\mathcal{U}}$ as follows:
\begin{equation}
\mathcal{Z}_{\mathcal{U}} = e(\{\boldsymbol{p}_n\}_{n \in \mathcal{U}}).
\end{equation}
These representations, along with a learned mask token $\boldsymbol{z}_m$, are used by a decoder $d$ to get the reconstructed tokens $\{\hat{\boldsymbol{p}}_n\}$ as follows:
\begin{equation}
    \{\hat{\boldsymbol{p}}_n\}=d(\mathcal{Z}_{\mathcal{U}}, \boldsymbol{z}_m).
\end{equation}
The learning objective is computed over the masked positions on the reconstructed tokens as follows:
\begin{equation}
\mathcal{L}_{\text{UMR}} = \text{RecL}(\{\hat{\boldsymbol{p}}_n\}_{n \in \mathcal{M}}, \{\boldsymbol{p}_n\}_{n \in \mathcal{M}}),
\end{equation}
where $\text{RecL}$ is a reconstruction loss (e.g., mean average error or mean square error) and $\mathcal{L}_{\text{UMR}}$ is the resulting uni-modal reconstruction loss (i.e., reconstruction loss where input and output are of the same modality).

The \ac{CSMAE} model~\cite{hackstein2024exploring} extends this formulation by introducing two adaptations of \acp{MAE} to enable cross-modal representation learning:
i) extending the encoder into sensor-specific encoders and a cross-sensor encoder; and
ii) extending the learning objective with latent similarity preservation and cross-modal reconstruction.
To accomplish this, first two independent masks $\mathcal{M}^\mathcal{X}$ and $\mathcal{M}^\mathcal{Y}$ and their inverse $\mathcal{U}^\mathcal{X}$ and $\mathcal{U}^\mathcal{Y}$ are created.
These are used on the input images, as described above, to get a set of tokens for each image modality.
Then, the sensor-specific encoders, which use one \ac{ViT} encoder per modality, are applied to the unmasked patches.
Here, each \ac{ViT} encoder ($e_{MS}^\mathcal{X}$ or $e_{MS}^\mathcal{Y}$) has its own set of parameters, enabling the accurate modeling of sensor-specific image characteristics.
Then the cross-sensor encoder $e_{CS}$ (a \ac{ViT} encoder that uses shared weights among all modalities) is applied on the output of the sensor-specific encoder.
It aligns the latent representations of the different modalities into a shared embedding space by processing the output of the sensor-specific encoder as follows:
\begin{align}
    \mathcal{Z}_\mathcal{U}^\mathcal{X} &= e_{CS}(e_{MS}^\mathcal{X}(\{\boldsymbol{p}^\mathcal{X}_n\}_{n \in \mathcal{U}^\mathcal{X}})),\\
    \mathcal{Z}_\mathcal{U}^\mathcal{Y} &= e_{CS}(e_{MS}^\mathcal{Y}(\{\boldsymbol{p}^\mathcal{Y}_n\}_{n \in \mathcal{U}^\mathcal{Y}})).
\end{align}

To enable cross-modal alignment, the \ac{CSMAE} model is trained using two cross-modal learning objectives:
i) cross-modal reconstruction; and
ii) latent similarity preservation.
For cross-modal reconstruction, two sensor-specific decoders $d^\mathcal{X}$ and $d^\mathcal{Y}$ are employed.
Each decoder takes as input the features of one modality ($\mathcal{Z}_\mathcal{U}^\mathcal{X}$ or $\mathcal{Z}_\mathcal{U}^\mathcal{Y}$) as well as a sensor-specific mask token ($\boldsymbol{z}^\mathcal{X}_m$ or $\boldsymbol{z}^\mathcal{Y}_m$) to reconstruct the masked patches of the other modality as follows:
\begin{align}
    \{\hat{\boldsymbol{p}}^\mathcal{Y}_n\}&=d^\mathcal{Y}(\mathcal{Z}_\mathcal{U}^\mathcal{X}, \boldsymbol{z}^\mathcal{Y}_m),\\
    \{\hat{\boldsymbol{p}}^\mathcal{X}_n\}&=d^\mathcal{X}(\mathcal{Z}_\mathcal{U}^\mathcal{Y}, \boldsymbol{z}^\mathcal{X}_m).
\end{align}
The cross-modal reconstruction loss $\mathcal{L}_{\text{CMR}}$ is then calculated using the same reconstruction loss as the uni-modal reconstruction as follows:
\begin{align}
\mathcal{L}_{\text{CMR}} &= \text{RecL}(\{\hat{\boldsymbol{p}}^\mathcal{Y}_n\}_{n \in \mathcal{M}^\mathcal{X}}, \{\boldsymbol{p}^\mathcal{Y}_n\}_{n \in \mathcal{M}^\mathcal{X}})\nonumber\\
&+ \text{RecL}(\{\hat{\boldsymbol{p}}^\mathcal{X}_n\}_{n \in \mathcal{M}^\mathcal{Y}}, \{\boldsymbol{p}^\mathcal{X}_n\}_{n \in \mathcal{M}^\mathcal{Y}}).
\end{align}
Note that here, for the loss calculation of one modality, the mask of the other modality is used, as this mask is also used to calculate the latent features.

For preservation of latent similarity, the mutual information loss $\mathcal{L}_\text{MI}$~\cite{Sumbul2022MIM}, which is based on contrastive learning with normalized temperature-scaled cross-entropy~\cite{Bachman2019NTXent}, is used. 
For a batch of $|\mathcal{B}|$ image pairs, let $\boldsymbol{c}_i^\mathcal{X}, \boldsymbol{c}_i^\mathcal{Y}$ be projected representations (e.g., projected \texttt{[CLS]} tokens) that should be aligned in the latent space.
Then, using a similarity function $\text{sim}(\cdot, \cdot)$ (e.g., cosine similarity), a temperature parameter $\tau$ and the indicator function $\mathds{1}$, $\mathcal{L}_\text{MI}$ can be defined as follows:
\begin{align}
\ell^i(\mathcal{X},\mathcal{Y}) &= -\text{log}\left(\frac{\exp{(\text{sim}(\boldsymbol{c}_i^\mathcal{X}, \boldsymbol{c}_i^\mathcal{Y})/\tau})}{\sum_{q \in \mathcal{B}}\mathds{1}_{[q \neq i]}\exp{(\text{sim}(\boldsymbol{c}_i^\mathcal{X}, \boldsymbol{c}_q^\mathcal{Y})/\tau})}\right),\\
\mathcal{L}_{\text{MI}}(\mathcal{B}) &= \frac{1}{2|\mathcal{B}|}\sum_{i\in\mathcal{B}} \left(\ell^i(\mathcal{X},\mathcal{Y}) + \ell^i(\mathcal{Y},\mathcal{X})\right).
\end{align}

Additionally, the same learning objective $\mathcal{L}_{\text{UMR}}$ as in single-modal \acp{MAE} is applied on both modalities individually.

\subsection{The proposed CSMoE Model}\label{sec:csmoe}
\begin{figure*}
    \centering
    \includegraphics[width=0.98\linewidth]{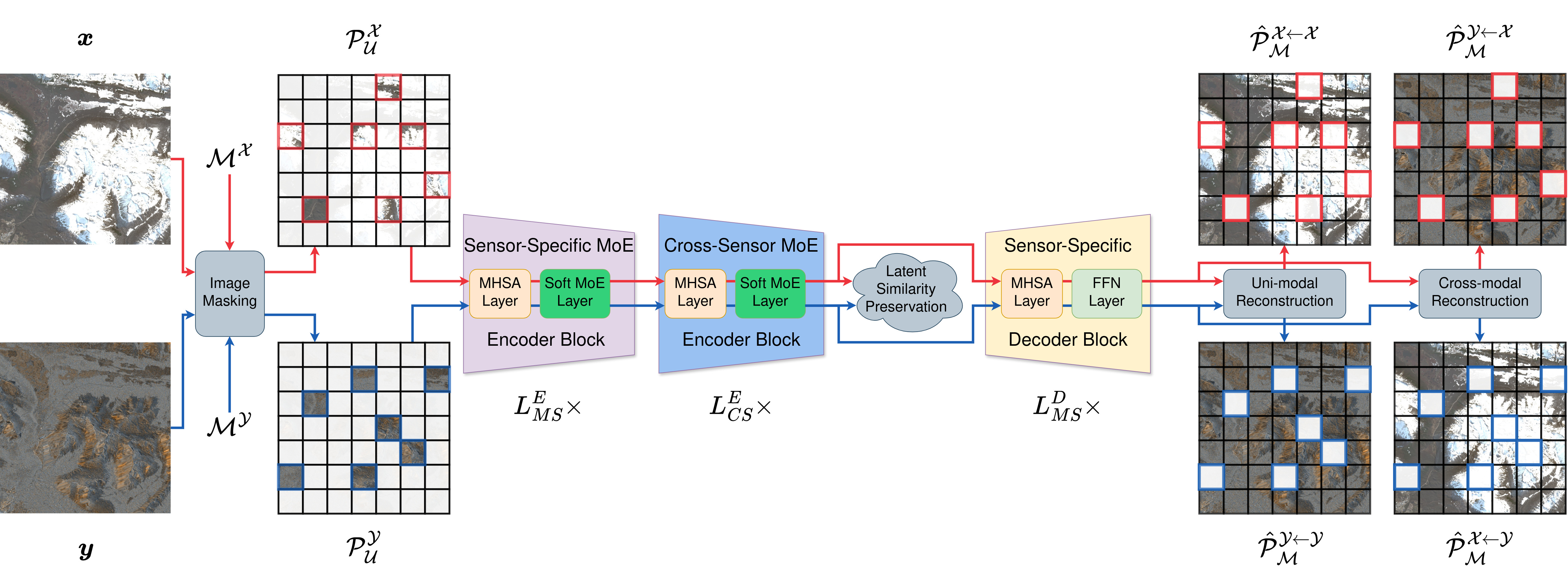}
    \caption{An illustration of the \ac{CSMoE} model for two image modalities. Each modality has one sensor-specific encoder consisting of $L_{MS}^E$ Transformer-\ac{MoE} layers and one sensor-specific decoder consisting of $L_{MS}^D$ Transformer layers. Additionally, all modalities share a cross-sensor encoder consisting of $L_{CS}^E$ Transformer-\ac{MoE} layers. Different colors indicate the processing pathways for each \ac{RS} image modality.}
    \label{fig:csmoe_arch_overview}
\end{figure*}

To simultaneously address the computational complexity limitations of masked autoencoder–based models while preserving cross-modal representation learning, we introduce a general adaptation for efficient multi-modal processing. 
Although the proposed adaptation can be applied to any transformer-based \ac{FM}, in this paper we apply the adaptation to \ac{CSMAE}, resulting in the \acf{CSMoE} model, the first compute-efficient multi-modal \ac{FM} for \ac{RS}.
The proposed \ac{CSMoE} model extends \ac{CSMAE} by integrating Soft \ac{MoE} mechanisms into the encoder components, which reduces the number of expert calls while maintaining both intra-modal and inter-modal characteristics. 
To achieve this, we adapt \ac{CSMAE} by: i) integrating expert routing; ii) adapting the encoder architecture; and iii) including regularizing training objectives. 
\Cref{fig:csmoe_arch_overview} shows an illustration of the \ac{CSMoE} model, while our adaptations are explained in detail in the following.

\begin{figure}
    \centering
    \includegraphics[trim={0 0 0 0},clip,width=0.98\linewidth]{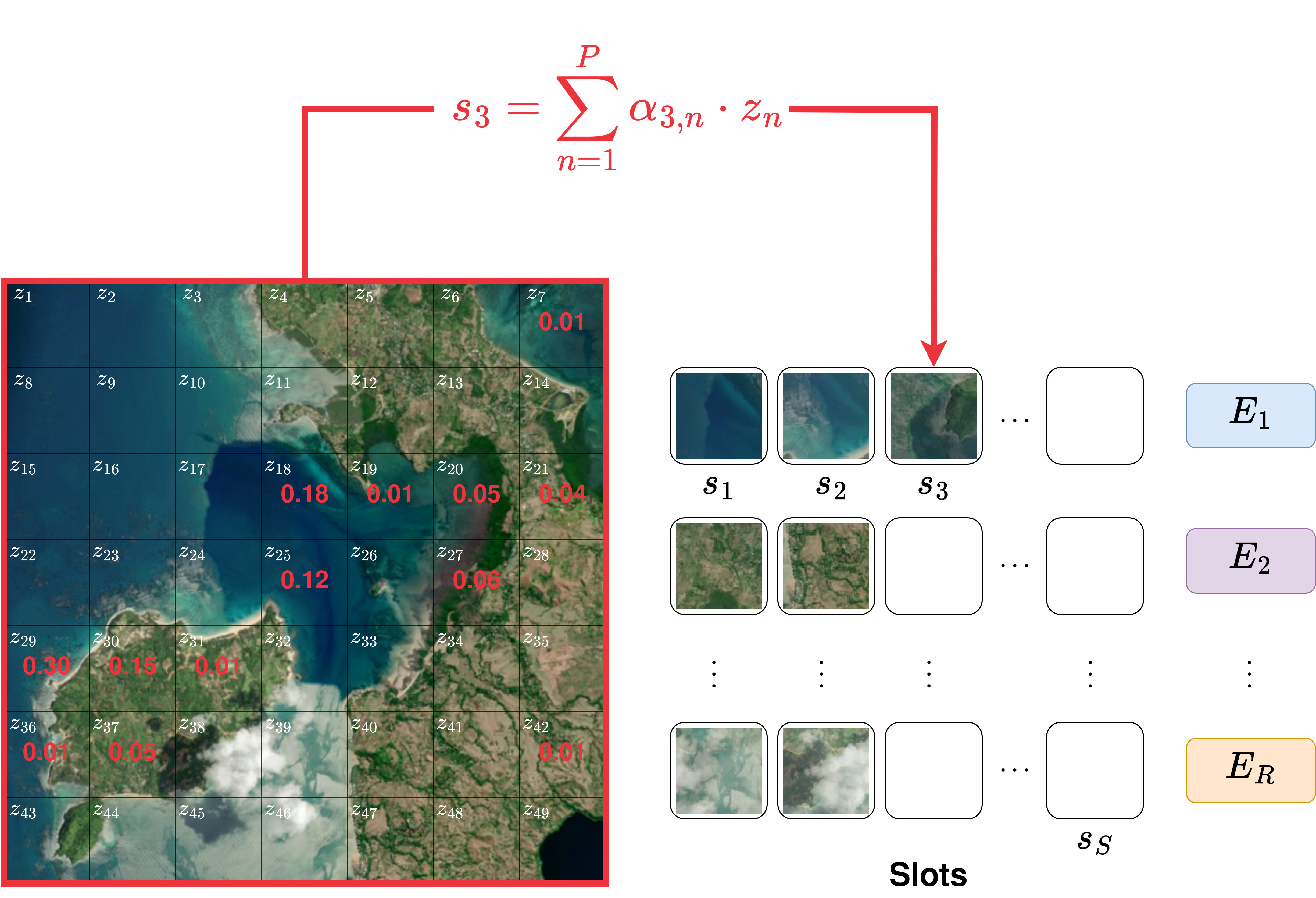}
    \caption{Qualitative example of the dispatch process. Slots ($s_1,\, ... \, s_S$) are formed as a weighted combination of the input features ($z_1,\,\dots\,z_{49}$ for $P=49$ patches) and assigned to experts ($E_1,\,\dots\,E_R$). The dispatch weights $\alpha_{3,n}$ for slot 3 are shown in red, features without annotation have a zero-weight. Note that in the model this process happens in the embedding space but for visualization it is shown in the image space.}
    \label{fig:softrsmoe}
\end{figure}
The proposed \ac{CSMoE} model leverages Soft \ac{MoE} to reduce computational complexity while maintaining model capacity through softly routed expert activation. 
Unlike sparse \acp{MoE}~\cite{shazeer2017moe} that route individual tokens to top-$k$ out of $R$ experts, Soft \ac{MoE} employs a two-stage routing mechanism based on $S$ intermediate representations called slots $\{\boldsymbol{s}_\vartheta\}_{\vartheta=1}^S$. 
As shown in \cref{fig:softrsmoe}, each slot summarizes a subset of input features through weighted aggregation, where slot–feature similarities are converted into dispatch weights $\alpha_{\vartheta,n}$ and combine weights $\hat{\alpha}_{\vartheta,n}$ as follows:
\begin{equation}
\alpha_{\vartheta,n} = \text{softmax}_n\left( \frac{\hat{s}_{\vartheta,n}}{\tau} \right), \quad
\hat{\alpha}_{\vartheta,n} = \text{softmax}_\vartheta(\hat{s}_{\vartheta,n}),
\end{equation}
where $\tau$ is a temperature parameter, $\text{softmax}_n$ and $\text{softmax}_\vartheta$ are softmax operations along the token and slot dimensions, respectively and $\{\hat{s}_{\vartheta,n}\}_{n=1}^P$ are slot embeddings (learned projections of the input features $\{z_n\}_{n=1}^P$) for expert $\vartheta$.
Each slot $s_\vartheta$ is formed as a weighted average of the input features as follows:
\begin{equation}
    s_\vartheta = \sum_{n=1}^P \alpha_{\vartheta,n} \cdot z_n.
\end{equation}
The slots are then processed by one assigned expert each and the final outputs are reconstructed through weighted aggregation.
This approach reduces expert calls from $k \cdot P$ to $S$, enabling efficient processing while preserving representational capacity.

The \ac{CSMoE} model incorporates Soft \ac{MoE} into both encoder components of \ac{CSMAE}.
All sensor-specific encoders employ an individual \ac{MoE} layer, where separate expert networks with different parameters are utilized for different image modalities, allowing accurate modeling of sensor-specific characteristics. 
Additionally, the cross-sensor encoder employs shared \ac{MoE} layers, where the same expert networks process features from all modalities to facilitate inter-modal pattern learning. 
Based on this, we define the \ac{CSMoE} model as follows:

\subsubsection{Sensor-Specific Encoding} Each image modality is processed through its dedicated encoder consisting of $L_\text{MS}^E$ Soft \ac{MoE} transformer encoder layers. 
For example, for modality $\mathcal{X}$, the unmasked tokens $\mathcal{P}_\mathcal{U}^\mathcal{X}$ are processed using its sensor-specific encoder $e_{MS}^\mathcal{X}$ as follows:
\begin{equation}
\mathcal{Z}_\text{MS}^\mathcal{X} = e_{MS}^\mathcal{X}(\mathcal{P}_\mathcal{U}^\mathcal{X}),
\end{equation}
where each encoder layer applies self-attention followed by Soft \ac{MoE} processing. 
For modality $\mathcal{Y}$, the processing is identical using its respective sensor-specific encoder $e_{MS}^\mathcal{Y}$.
These sensor-specific encoders allow each image modality to be processed by an independent set of experts, and thus to be modeled with its own parameters.
\chadded{Note that this design does not enforce a semantic specialization of the individual experts. 
The routing behavior that emerges during pretraining is analyzed in \cref{sec:expert_analysis}.}

\subsubsection{Cross-Sensor Encoding} The outputs from both sensor-specific encoders are processed through a shared cross-sensor encoder $e_{CS}$ consisting of $L_\text{CS}^E$ Soft \ac{MoE} transformer encoder layers:
\begin{align}
\mathcal{Z}_\text{CS}^\mathcal{X} &= e_{CS}(\mathcal{Z}_\text{MS}^\mathcal{X}) \\
\mathcal{Z}_\text{CS}^\mathcal{Y} &= e_{CS}(\mathcal{Z}_\text{MS}^\mathcal{Y}).
\end{align}
The shared Soft \ac{MoE} transformer encoder layers process the features of both image modalities with the same set of experts, and thus operate on a common representation space, facilitating inter-modal relationship modeling while maintaining computational efficiency.

\subsubsection{Sensor-Specific Decoding} For reconstruction, separate decoders $d_{MS}^j$ consisting of $L_\text{MS}^D$ standard transformer layers are employed.
The decoders process cross-sensor representations to reconstruct both uni-modal and cross-modal targets as follows:
\begin{equation}
\hat{\mathcal{P}}_\mathcal{M}^{j\leftarrow j^\prime} = \text{Linear}_{j\leftarrow j^\prime}(d_{MS}^j(\mathcal{Z}_\text{CS}^{j^\prime}, \boldsymbol{z}_m^j)),
\end{equation}
where $j, j^\prime \in \{\mathcal{X}, \mathcal{Y}\}$ and $j \leftarrow j^\prime$ denotes the reconstruction of the image modality $j$ using features of the image modality $j^\prime$. 
The decoders do not include \ac{MoE} components as they are removed after pretraining.

\subsubsection{Training Objective Extensions}
To regularize expert routing and ensure stable training, we extend the \ac{CSMAE} training objectives with two additional loss functions.
The slot repulsion loss $\mathcal{L}_\text{REP}$ encourages reduced correlation between slot embeddings as follows:
\begin{equation}
    \mathcal{L}_{\text{REP}} = \frac{1}{S(S-1)} \sum_{\vartheta=1}^{S} \sum_{\substack{\vartheta'=1 \\ \vartheta' \neq \vartheta}}^{S} \left( \langle s_{\vartheta}, s_{\vartheta'} \rangle \right)^{2},
\label{eq:rep}
\end{equation}
while the entropy loss $\mathcal{L}_\text{ENT}$ promotes balanced expert usage as follows:
\begin{equation}
    \mathcal{L}_\text{ENT} = -\frac{1}{SP}\sum_{\vartheta=1}^{S} \sum_{n=1}^{P} \alpha_{\vartheta,n} \log \left( \alpha_{\vartheta,n} + \varepsilon \right),
\end{equation}
where $\varepsilon$ is a small term for numerical stability.

The overall training objective combines the \ac{CSMAE} losses with regularization terms as follows:
\begin{equation}
    \mathcal{L}_\text{total} = \mathcal{L}_\text{UMR} + \mathcal{L}_\text{CMR} + \mathcal{L}_\text{MI} + \lambda \cdot \mathcal{L}_\text{REP} + \gamma \cdot \mathcal{L}_\text{ENT},
\end{equation}
where $\lambda$ and $\gamma$ are scaling parameters controlling the regularization influence. 
In this way, the characterization of cross-modal \ac{RS} image representations is achieved by learning to reconstruct both intra-modal and inter-modal image content while maintaining computational efficiency through selective expert activation and promoting stable expert usage through regularized routing.

\subsection{A Thematic-Climatic Descriptor-Driven Sampling Strategy}
\label{sec:sampling}
The efficiency of \ac{FM}-training in \ac{RS} depends not only on model design choices, but also on training data selection.
The use of global datasets often introduces substantial redundancy, where vast amounts of data contribute little additional information. 
For instance, when training on global coverage, more than 70\% of the images may depict oceans, which are irrelevant for most land-focused applications. 
Even within land areas, extended homogeneous regions, such as the Amazon rainforest, the Siberian taiga, or the Sahara desert, dominate the dataset but provide limited additional information.

To address this issue, we propose a thematic-climatic descriptor-driven sampling strategy that samples a reduced but representative training set from a large-scale \ac{RS} data archive.
This strategy efficiently samples a training set with a wide range of geographic and thematic-climatic variability.
The strategy consists of two stages:
i) auxiliary descriptor generation, where each image is associated with climatic and thematic descriptors derived from global \ac{CZ} and thematic product datasets; and
ii) entropy-maximizing stratified sampling, where a genetic algorithm selects spatially dispersed and thematically-climatically balanced samples.

\subsubsection{Auxiliary Descriptor Generation}
\label{sec:class-assignment}

Let $\mathcal{D} = \{\varphi_i\}_{i=1}^{N}$ denote the set of bounding boxes (the set of coordinates that define the spatial extent) of the images within a large-scale \ac{RS} data archive $\mathcal{T}$ that consists of $N$ images or co-registered image tuples, where $\varphi_i$ contains the longitude and latitude coordinates of the bounding box of the $i$th image of $\mathcal{T}$.
We assume that the images are unlabeled, i.e., no thematic (e.g., \ac{LULC}) or climatic information is directly available.
In the first stage, each image is associated with a thematic and a climatic descriptor derived from global raster layers.
Specifically, each bounding box $\varphi_i$ is linked to: 
i) a \ac{CZ} raster $M_{\text{climate}}$ (e.g., Köppen--Geiger~\cite{koppengeiger}), which provides coarse climatic information; and 
ii) a thematic land-cover product $M_{\text{TP}}$ (e.g., ESA WorldCover~\cite{zanaga2022esaworldcover}), which captures land-use properties at finer scale.
This combination allows us to incorporate both climate and local land-cover information, two complementary dimensions of variability, into the sampling process.

In detail, each bounding box $\varphi_i$ is associated with a climate stratum $u_i$ and a thematic stratum $v_i$, depending on the overlay with the reference rasters.
The resulting set of tuples is therefore represented as $\mathcal{D'} = \{(\varphi_i, u_i, v_i)\}_{i=1}^{N'} \;|\; N' \leq N$, where only images covered by both rasters are retained. 
We refer to these thematic-climatic strata as descriptors since they are not manually labeled but derived from external data sources.
These descriptors are used for the subsequent stratified sampling stage by embedding thematic and climatic variability directly into the sampling process.
The algorithm of the descriptor generation stage is given in \cref{alg:assign_classes}.
\begin{algorithm}
\caption{Assign Thematic and Climatic Descriptors to Coordinates}
\label{alg:assign_classes}
\begin{algorithmic}[1]
\Require Set of coordinates $\mathcal{D} = \{\varphi_i\}_{i=1}^N$
\Require Climate map $M_{\text{climate}}$ with bounds $M_{\text{climate}}^B$ 
\Require Thematic product map $M_{\text{TP}}$ with bounds $M_{\text{TP}}^B$
\Ensure Descriptors for each coordinate: climate stratum $u_i$ and thematic product stratum $v_i$

\State Initialize empty annotated coordinates $\mathcal{D'}= \{\}$
\For{each coordinate $\varphi_i \in \mathcal{D}$}
    \If{$\varphi_i \not\subset M_{\text{climate}}^B$ or $\varphi_i \not\subset M_{\text{TP}}^B$}
        \State Continue to next coordinate
    \EndIf

    \State $u_i \gets M_{\text{climate}}[\varphi_i]$
    \State $v_i \gets M_{\text{TP}}[\varphi_i]$

    \State Add $(\varphi_i, u_i, v_i)$ to $\mathcal{D'}$
\EndFor
\State \Return Set of pseudo-annotated tuples $\mathcal{D'}$
\end{algorithmic}
\end{algorithm}

\subsubsection{Entropy-Maximizing Stratified Sampling}
\label{sec:entropy-sampling}

Given the set of tuples $\mathcal{D'}$ containing locations and descriptors, the entropy-maximizing stratified sampling aims to construct a reduced training set $\mathcal{D}^\star$ that preserves the spatial and thematic-climatic variability of $\mathcal{T}$ while minimizing redundancies among the selected samples.
To this end, we stratify $\mathcal{D'}$ by joined thematic and climatic descriptors and then sample within each joined stratum using an entropy-driven sampling process.
Specifically, for each pair $(u, v)$ of climatic and thematic strata, we define:
\begin{equation}
\mathcal{S}_{u,v} = \left\{ (\varphi_i, u_i, v_i) \in \mathcal{D'} \;\middle|\; u_i = u,\; v_i = v \right\}.
\end{equation}
While some joined strata are small and can be fully retained (i.e., the size of the joined stratum is smaller than a user-defined target sample count per stratum $N_s$), others are extensive and can be sampled from to reduce redundancy.
In contrast to random sampling, we use the spatial dispersion (distance) of selected samples within each joined stratum as a criterion for sampling.
To this end, we employ a genetic algorithm that iteratively optimizes a subset of samples within $\mathcal{S}_{u,v}$.
Each candidate solution $\{\mathcal{S}_{u,v} \mid \boldsymbol{b}\}$ is a set of samples, where $\boldsymbol{b}$ is a binary mask indicating which samples of the joined stratum are selected for the candidate solution.
The fitness of the candidate solution is measured by an entropy-based score that uses the pairwise Haversine distances. 
This encourages the selection of samples that are representative and spatially diverse. 
Standard evolutionary operators are used to optimize candidate solutions.
Additionally, to maintain the target sample counts per joined stratum $N_s$, we use an adaptive constraint mechanism: if a candidate solution exceeds $110\%$ of the target sample count, it is randomly pruned; if it falls below $90\%$ it is randomly augmented.  

After convergence, or if a given compute budget is reached, the candidate solution with the highest fitness is selected.
The process is repeated independently across all joined strata, and the union of candidate solutions with the highest respective fitness defines the training set $\mathcal{D}^\star$. 
This strategy ensures that the resulting training set is spatially well-dispersed, thematically and climatically balanced, and substantially less redundant than the original archive.
The algorithm of the optimization stage is given in \cref{alg:genetic_entropy_sampling}.  

\begin{algorithm}[ht]
\caption{Entropy-Based Subsampling via Genetic Optimization}
\label{alg:genetic_entropy_sampling}
\begin{algorithmic}[1]
\Require Set of location-descriptor tuples $\mathcal{D'}$
\Require Number of samples per joined stratum $N_s$, number of iterations $T$, population size $N_p$, climate zones $C_u$, thematic product classes $C_v$
\Require Mutation rate $r_m$, crossover rate $r_c$
\Ensure Optimized stratified sample subset $\mathcal{D}^*$ with high spatial entropy
\State Initialize empty sample subset $\mathcal{D}^\star = \{\}$
\ForAll{climatic strata $u \in C_u$}
    \ForAll{thematic strata $v \in C_v$}
        \State Extract subset $\mathcal{S}_{u,v} \subseteq \mathcal{D'}$ where $u_i=u$, $v_i=v$
        \State $n_g \leftarrow |\mathcal{S}_{u,v}|$
        \If{$n_g \leq N_s$}
            \State Add all elements of $\mathcal{S}_{u,v}$ to $\mathcal{D}^\star$
        \Else
            \State Define objective function \vfill
            $f_g(\boldsymbol{b}) \leftarrow \text{Entropy}(\{\mathcal{S}_{u,v} \mid \boldsymbol{b}\})$
            \State Run genetic algorithm \vfill
            $\boldsymbol{b}^\star \leftarrow \text{GA}(f_g, n_g, T, N_p, N_s, r_m, r_c)$
            \State $\mathcal{S}_{u,v}^\star \leftarrow \{\mathcal{S}_{u,v} \mid \boldsymbol{b}^\star\}$
            \State $\mathcal{D}^\star \leftarrow \mathcal{D}^\star\, \cup\, \mathcal{S}_{u,v}^\star$
        \EndIf
    \EndFor
\EndFor
\State \Return $\mathcal{D}^\star$
\end{algorithmic}
\end{algorithm}
\newpagetoggle

\section{Datasets and Experimental Setup}\label{sec:experiment}
\chadded{This section describes the pretraining dataset construction, the downstream tasks used for evaluation, and the experimental setup applied throughout the paper.}

\subsection{Efficient Sampling of Large-Scale Pretraining Data}
We pretrained on a dataset that we constructed based on \ac{MTC}~\cite{francis2024major}.
We selected \ac{MTC} as our large-scale \ac{RS} data archive since it is publicly available, contains globally distributed samples, and has already been adopted in recent studies, indicating its potential as an emerging benchmark for global-scale representation learning.
This makes it a suitable and reproducible foundation for our sampling strategy.
To construct our pretraining dataset, denoted as \ac{mTC}, we used our proposed thematic-climatic descriptor-driven sampling strategy on \ac{MTC}.
Additionally, to evaluate the effect of using the entropy-maximizing stratified sampling, we created a second dataset, denoted as \ac{rTC}, which was created using the same thematic-climatic stratified sampling but without applying the genetic algorithm to increase the spatial diversity.
Both datasets include only tiles with matching \ac{S1} \ac{SAR} and \ac{S2} multispectral image pairs. 
For auxiliary descriptor generation, we utilized the Köppen-Geiger climate classification map~\cite{koppengeiger} for \ac{CZ} information and ESA WorldCover~\cite{zanaga2022esaworldcover} as the thematic product.
We associated each tile in \ac{MTC} with a Köppen-Geiger class and ESA WorldCover class.
For computational reasons, we only used the center location of each tile for this.
Based on the association, we used our entropy-maximizing stratified sampling for \ac{mTC} and a random selection for \ac{rTC}.
Both sets contained $N_s \approx 100$ samples per joined stratum.
For the sampling of \ac{mTC}, the genetic algorithm was run for $T = 2\,500$ iterations with population size $N_p=10$, crossover rate $r_c = 0.5$, and mutation rate $r_m =\frac{N_s}{n_g \cdot 25}$, where $n_g$ is the total number of samples in the joined stratum.

\Cref{tab:dataset_stats_MTC} shows that both datasets have similar class balance and distance between tiles, but \ac{mTC} contains a significantly greater spatial diversity within each class.
\cref{fig:samples_log} visualizes the distribution of samples with respect to their location.
As one can see from the figure, large homogeneous regions (e.g.,  the Amazon forest, the Sahara desert, or Siberia) and small islands exhibit significantly improved spatial diversity when entropy-maximizing stratified sampling is applied to the dataset.
Following~\cite{clasen2025refinedbigearthnet}, we split each tile of \ac{mTC} into patches of size $120\times120$ pixels, discarding those that are too small (due to non-integer multiples of tile-patch size relations) or contain invalid pixels (see \cref{fig:splitting}).
This yielded a total of 1\,055\,080 valid training patches, of which we randomly selected ~5\% to track for validation loss during the training.

\begin{table}
\centering
\renewcommand{\arraystretch}{1}
\setlength\tabcolsep{3pt}
\caption{Statistics on the full \ac{MTC} dataset, a \chadded{random} stratified subset \ac{rTC} and our entropy-\chadded{maximized} stratified subset \ac{mTC}.}
\label{tab:dataset_stats_MTC}
\begin{tabular}{lrrr}
\toprule
Statistic & \ac{MTC} & \ac{rTC} & \ac{mTC} \\ 
\midrule
Number of tiles                                & 1\,302\,691     &  18\,785     & 18\,846     \\
Tiles per \ac{CZ}                        &     43\,423     &      626     &     628     \\
Tiles per LULC class                           &    118\,426     &   1\,707     &  1\,713     \\
\makecell[l]{Tiles per combination\\($N_s>90$)}& 7\,434 $\sigma$ 16\,395 & 98 $\sigma$ 4 & 100 $\sigma$ 2 \\
Average Distance                               &      6\,737\,km &   8\,192\,km &  8\,648\,km \\
Min. distance                                  &          10\,km &       32\,km &      36\,km \\
\makecell[l]{Distance within $S_{u,v}$}        &      2\,748\,km &   1\,811\,km &  5\,106\,km \\
\bottomrule
\end{tabular}
\end{table}




\begin{figure*}[htbp]
    \centering
    \begin{subfigure}[b]{0.49\textwidth}
        \centering
        \begin{tikzpicture}
            \node[anchor=south west,inner sep=0] (image) at (0,0) {
                \includegraphics[width=\textwidth]{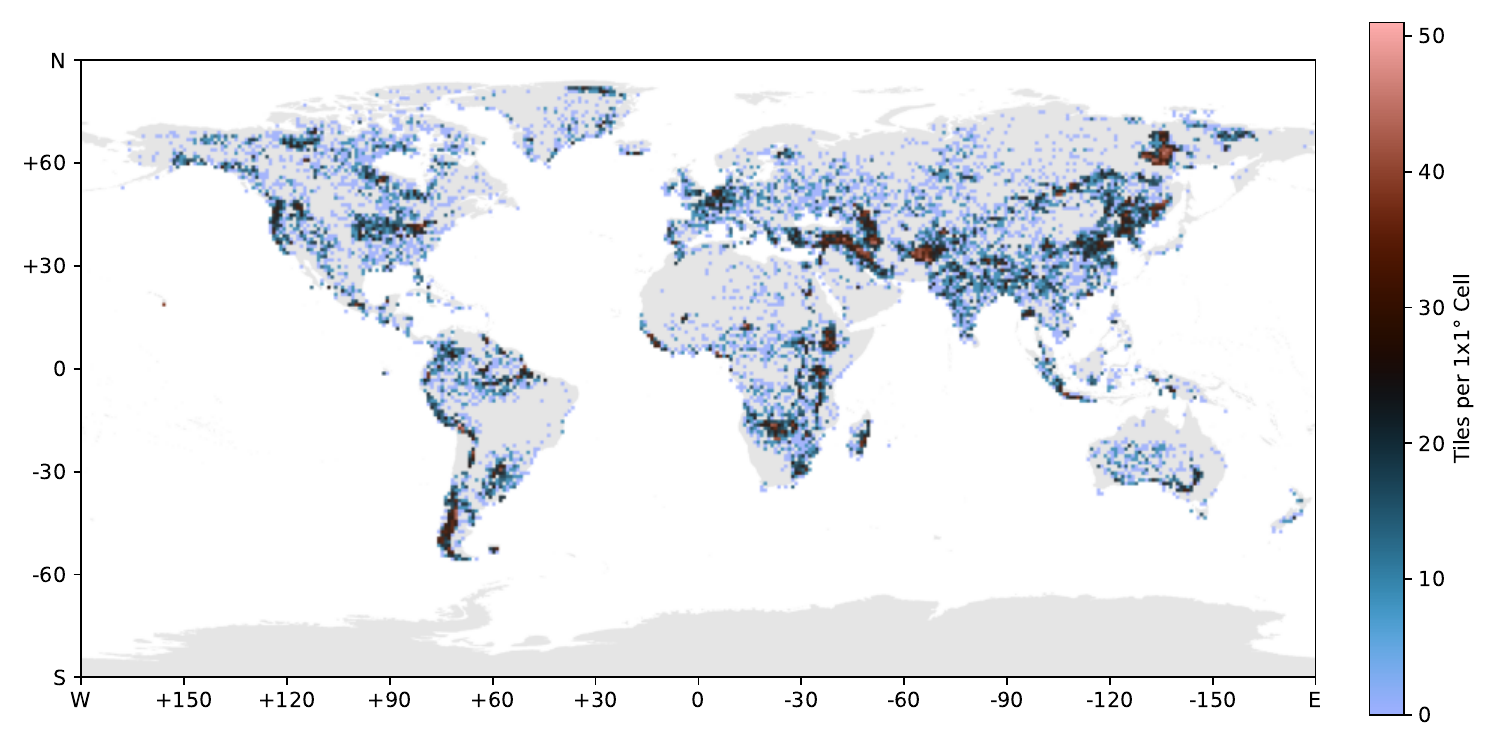}
            };
            \begin{scope}[x={(image.south east)}, y={(image.north west)}]
                \draw[red, thick] (0.35, 0.45) circle [x radius=0.05, y radius=0.08]; 
                \draw[red, thick] (0.49, 0.60) circle [x radius=0.07, y radius=0.08]; 
                \draw[red, thick] (0.65, 0.77) circle [x radius=0.12, y radius=0.08]; 
                \draw[red] (0.38, 0.26) circle [x radius=0.005, y radius=0.01]; 
                \draw[red] (0.625, 0.28) circle [x radius=0.005, y radius=0.01]; 
                \begin{scope}[x={(image.south east)}, y={(image.north west)}]  
                    \begin{scope}[shift={(0.79,0.40)}, rotate=320]
                        \draw[red, thick] (0,0) ellipse [x radius=0.045, y radius=0.05];
                    \end{scope}
                \end{scope}
            \end{scope}
        \end{tikzpicture}
        \caption{}
        \label{fig:samples_log_rand}
    \end{subfigure}
    \hfill
    \begin{subfigure}[b]{0.49\textwidth}
        \centering
        \includegraphics[width=\textwidth]{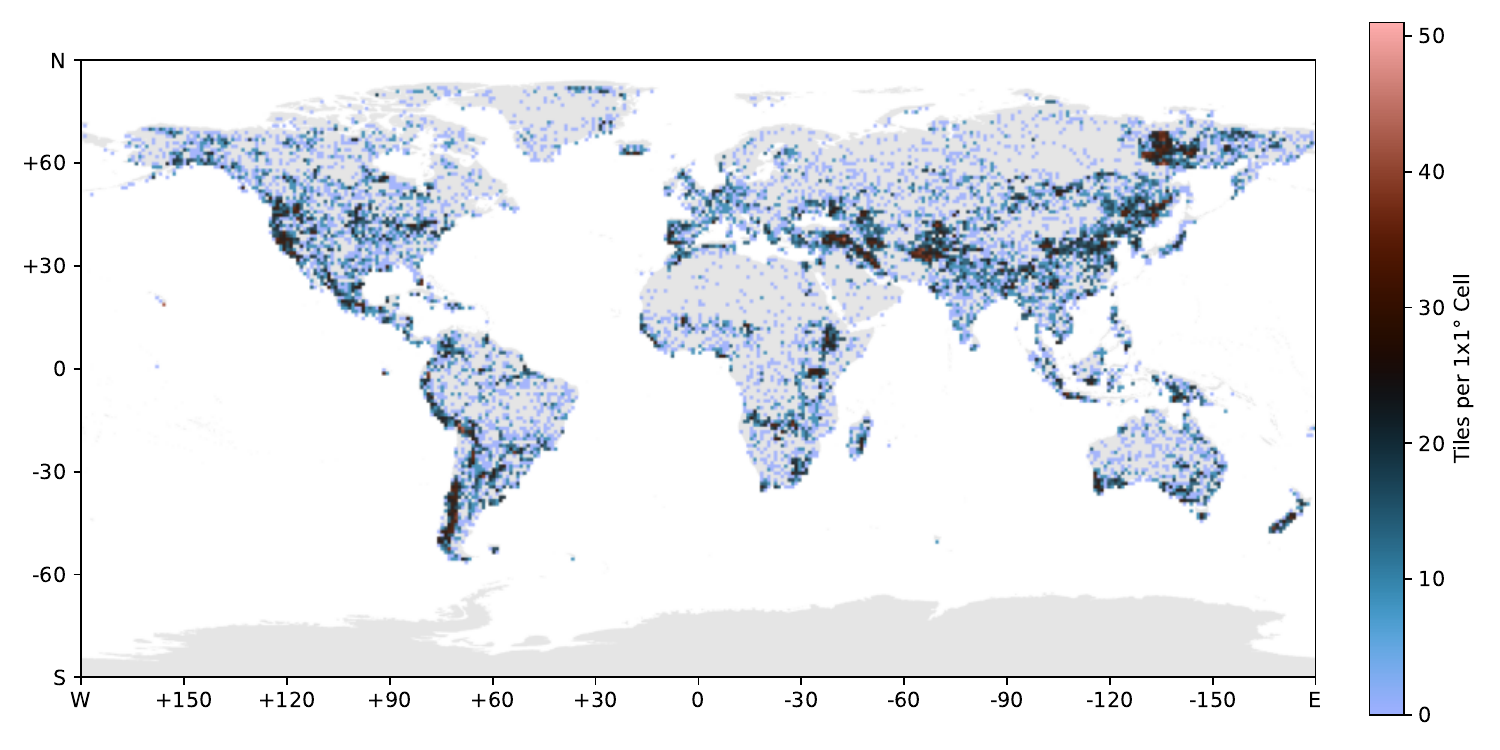}
        \caption{}
        \label{fig:samples_log_sampled}
    \end{subfigure}
    
    \caption{Tile density per $1^\circ \times 1^\circ$ square (a) in random sampling (\ac{rTC}); and (b) using our entropy-maximizing stratified sampling (\ac{mTC}). Red circles mark regions that have a significantly lower spatial diversity of samples in \ac{rTC} compared to \ac{mTC}, e.g., the Amazon forest, the Sahara desert, or Siberia.}
    \label{fig:samples_log}
\end{figure*}
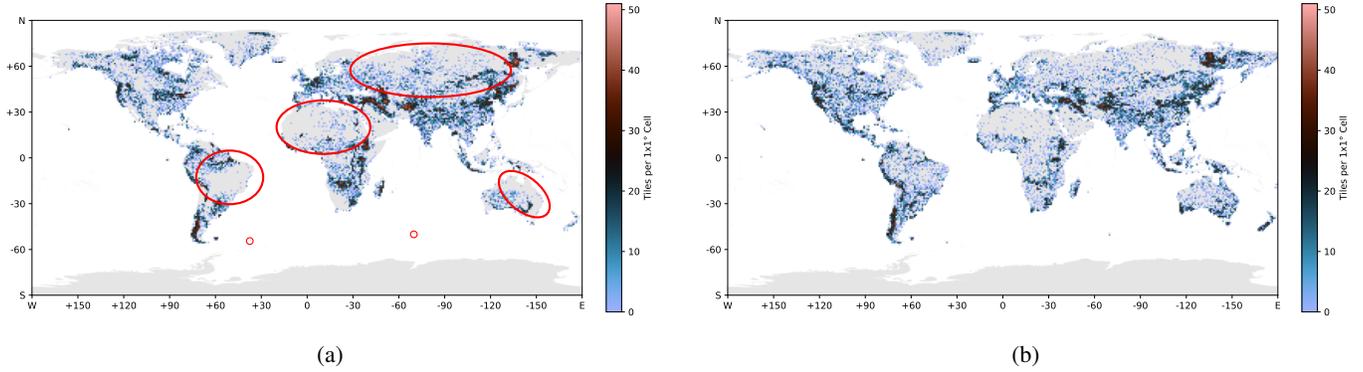
\begin{figure}
    \centering
    \begin{subfigure}[b]{0.24\textwidth}
        \centering
        \includegraphics[trim={0 1200 0 100},clip,width=\textwidth]{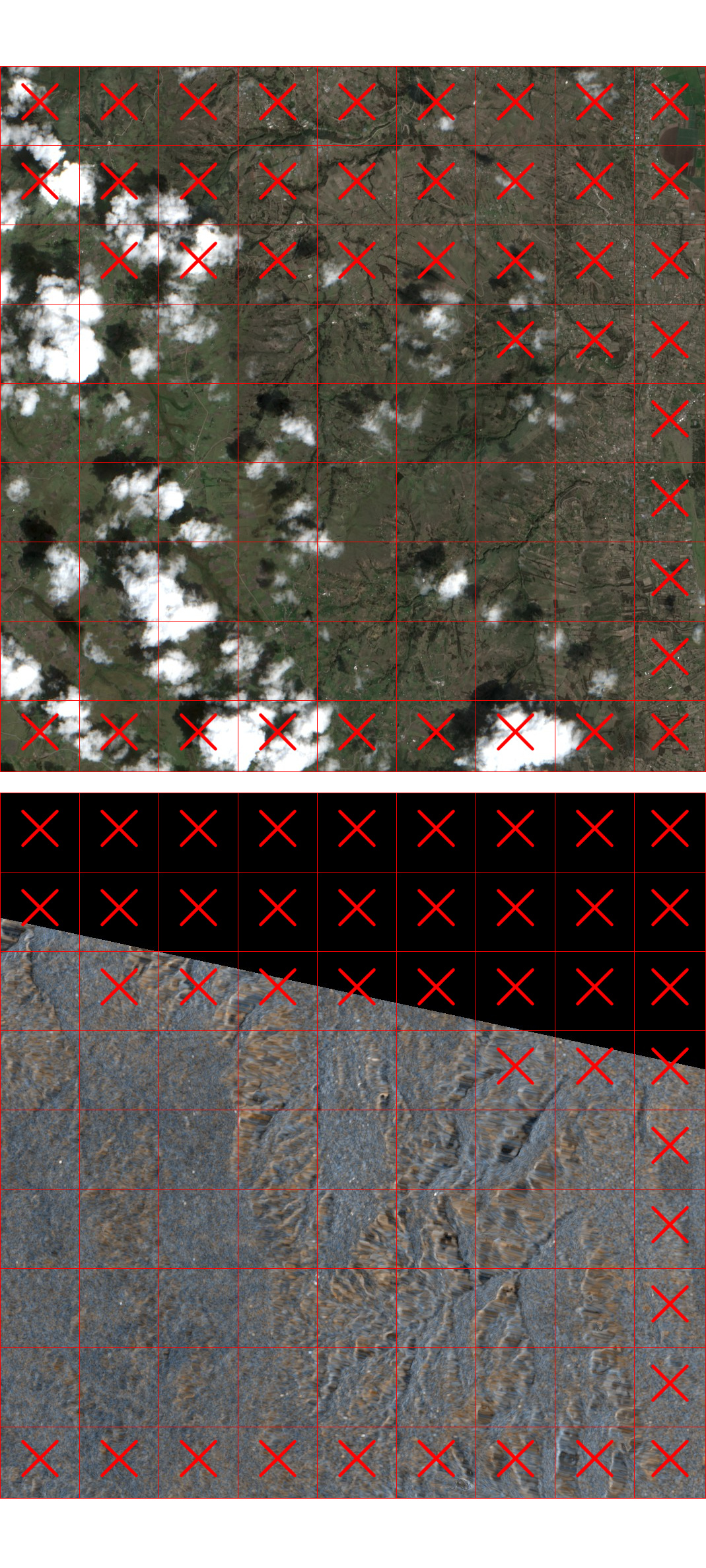}
        \caption{}
        \label{fig:splitting-S2}
    \end{subfigure}
    \hfill
    \begin{subfigure}[b]{0.24\textwidth}
        \centering
        \includegraphics[trim={0 100 0 1200},clip,width=\textwidth]{figures/images/MTomSplitting_10D_877L.drawio.95.jpg}
        \caption{}
        \label{fig:splitting-S1}
    \end{subfigure}
    
    \caption{Result of splitting and filtering a \ac{mTC} tile into patches: (a) True color representation of the Sentinel-2 tile; (b) False colour composites of the Sentinel-1 tile in decibel-scale with red channel as VV, green channel as VV+VH, and blue channel as VH. Patches marked with a red \textcolor{red}{$\times$} are dropped due to invalid pixels or incompatible size.}
    \label{fig:splitting}
\end{figure}
\begin{figure*}[ht!]
\centering

\begin{subfigure}{\textwidth}
  \centering
  \foreach \idx in {6776,6795,6798,6801,6812,6813}{%
    \begin{minipage}[t]{0.155\textwidth}
      \includegraphics[width=\linewidth]{figures/images/examples_sampling/random/random_idx_\idx.png}
    \end{minipage}%
    \hspace{1pt}%
  }\vspace{1pt}
  
  \foreach \idx in {6815,6818,6824,6826,6828,6836}{%
    \begin{minipage}[t]{0.155\textwidth}
      \includegraphics[width=\linewidth]{figures/images/examples_sampling/random/random_idx_\idx.png}
    \end{minipage}%
    \hspace{1pt}%
  }
  \caption*{(a)}
\end{subfigure}

\vspace{4mm}

\begin{subfigure}{\textwidth}
  \centering
  \foreach \idx in {6776,6795,6798,6801,6812,6813}{%
    \begin{minipage}[t]{0.155\textwidth}
      \includegraphics[width=\linewidth]{figures/images/examples_sampling/sampling/sampled_idx_\idx.png}
    \end{minipage}%
    \hspace{1pt}%
  }\vspace{1pt}
  
  \foreach \idx in {6815,6818,6824,6826,6828,6836}{%
    \begin{minipage}[t]{0.155\textwidth}
      \includegraphics[width=\linewidth]{figures/images/examples_sampling/sampling/sampled_idx_\idx.png}
    \end{minipage}%
    \hspace{1pt}%
  }
  \caption*{(b)}
\end{subfigure}

\caption{Qualitative comparison of (a) random stratified sampling and (b) entropy-maximizing stratified sampling across 12 examples each. All samples are from the subset of the respective dataset which is associated with \ac{CZ} \enquote{Tropical, rainforest} and \ac{LULC} \enquote{Cropland}.}
\label{fig:sampling_comparison}
\end{figure*}

\subsection{Description of Downstream Tasks}
We evaluated our pretrained \ac{CSMoE} model on four scene classification and two semantic segmentation tasks from the geobench benchmark collection~\cite{lacoste2023geobench} as well as on uni-modal and cross-modal \ac{CBIR} on the BigEarthNet-v2 benchmark dataset~\cite{clasen2025refinedbigearthnet}.
The datasets for scene classification were the following:
\begin{itemize}
    \item \textbf{m-bigearthnet} is a multi-label \ac{LULC} classification dataset covering ten European countries. It contains 22\,000 \ac{S2} images, each labeled with one or more of the 43 CLC2018~\cite{clcbook} classes.
    \item \textbf{m-brick-kiln} is a binary scene classification dataset for brick kiln detection in Bangladesh. It contains 17\,061 \ac{S2} images that are labeled with brick-kiln or no-brick-kiln.
    \item \textbf{m-so2sat} is a multi-class \ac{CZ} classification dataset. It contains 21\,964 \ac{S1} and \ac{S2} images from 42 globally distributed cities and is associated with one of 17 local \acp{CZ}.
    \item \textbf{m-eurosat} is a multi-class \ac{LULC} classification dataset covering Europe. It contains 4\,000 \ac{S2} images, each labeled with one of 13 classes.
\end{itemize}
For semantic segmentation, we used the following datasets:
\begin{itemize}
    \item \textbf{m-cashew-plant} is a cashew plantation segmentation dataset from 1\,800 \ac{S2} images obtained over Benin. Each pixel is labeled with one of seven classes.
    \item \textbf{m-SA-crop-type} is a crop-type segmentation dataset from \ac{S2} images. It contains 5\,000 images from Brandenburg, Germany, and Cape Town, South Africa. Each pixel is associated with one of ten crop type labels.
\end{itemize}
For \ac{CBIR}, we conducted experiments on the BigEarthNet-v2 benchmark dataset~\cite{clasen2025refinedbigearthnet}. 
We used the following sets of image pairs:
\begin{itemize}
    \item \textbf{BENv2-14k} consists of 13\,683 BigEarthNet-v2 image pairs acquired over Serbia during the summer months. Each \ac{S1}-\ac{S2} pair is labeled with one or more classes from the 19-class nomenclature introduced in~\cite{sumbul2021bigearthnet}.
    \item \textbf{BENv2-243k} consists of 243\,130 BigEarthNet-v2 image pairs acquired over the ten European countries during the summer and autumn months. Each \ac{S1}-\ac{S2} pair is labeled with one or more classes from the same 19-class nomenclature.
\end{itemize}

We separately stacked: i) the VV and VH bands of \ac{S1} images (if applicable); and ii) the \ac{S2} bands associated with 10m and 20m spatial resolution, while nearest-neighbor interpolation was applied to the 20m bands.
For the datasets in the geobench benchmark collection, we followed the proposed train/validation/test-split from~\cite{lacoste2023geobench}.
For the \ac{CBIR} experiments, the validation split as proposed in~\cite{clasen2025refinedbigearthnet} of the respective set of images was used to select query images, while images were retrieved from the test split.
We evaluated two scenarios for \ac{CBIR}: i) uni-modal \ac{CBIR}, where query images and retrieved images belong to the same image modality; and ii) cross-modal \ac{CBIR}, where query images were selected from one image modality and retrieved from the other image modality.

\subsection{Experimental Setup}
We trained our \ac{CSMoE} model in four variants with different patch sizes $\rho \in \{32, 28, 16, 14\}$.
If not noted differently, the \ac{CSMoE} model was trained for 150 epochs on \ac{mTC} with a mini-batch size of \SIrange{256}{512}{}, depending on the memory requirements of the model, and an image size of $224\times224$ pixels.
All model variants used four modality-specific encoder layers ($L_{MS}^E = 4$), two cross-sensor encoder layers ($L_{CS}^E = 2$), four modality-specific decoder layers ($L_{MS}^D = 4$), and eight experts per layer ($S=8$).
The embedding dimension was set to 768 for the encoder and 256 for the decoder.
During pretraining, we followed \cite{hackstein2024exploring} and set both the masking ratio and the temperature of the $\mathcal{L}_\text{MI}$ to 0.5.
The AdamW optimizer with learning rate $10^{-4}$ and a cosine annealing learning rate schedule with linear warm-up was utilized.
All the experiments were conducted on NVIDIA $4\times$A100 or $4\times$H200 GPUs.

We carried out three different kinds of experiments to: 
i) perform a sensitivity analysis with respect to different variants of the \ac{CSMoE} model;
ii) compare the \ac{CSMoE} model variants with other \acp{FM} in terms of their scene classification and semantic segmentation performance relative to their computational complexity; 
and iii) compare the \ac{CSMoE} model variants with the baseline \ac{CSMAE}~\cite{hackstein2024exploring} model in terms of their uni-modal and cross-modal \ac{CBIR} performance.
For the sensitivity analysis, we varied the: 1) patch size $\rho$; 2) strategies of constructing the classification token; 3) the number of pretraining epochs; \chadded{and 4) the sampling strategy used to construct the pretraining dataset, comparing our thematic-climatic descriptor-driven strategy against random sampling within the same stratified framework.}
For the comparison with other \ac{RS} \acp{FM}, we compared the \ac{CSMoE} variants with: Prithvi~V2~\cite{szwarcman2024prithvi} in the 300 million and 600 million parameter versions;
Satlas~\cite{bastani2023satlaspretrain} in the swin-base configuration; 
DOFA~\cite{xiong2024dofa} in the base configuration; 
TerraMind~\cite{jakubik2025terramind} in the base configuration; and
\ac{CSMAE}~\cite{hackstein2024exploring} in the SECD configuration.
For all models except for \ac{CSMAE}, we used the implementations and checkpoints provided in TerraTorch~\cite{gomes2025terratorch}.
For \ac{CSMAE}, we used the implementation provided in \cite{hackstein2024exploring} and trained one model on BigEarthNet-v2~\cite{clasen2025refinedbigearthnet} and one on \ac{mTC}.
We used the model trained on BigEarthNet-v2 for the scene classification and semantic segmentation following~\cite{hackstein2024exploring}, and the model trained on \ac{mTC} for uni-modal and cross-modal \ac{CBIR} to avoid evaluating based on training data bias.

For the scene classification and semantic segmentation tasks, we considered probing as our downstream scenario, where we froze the backbone of the \ac{FM} and only trained a linear layer and a UPerNet~\cite{xiao2018upernet} decoder for scene classification and semantic segmentation, respectively, for 50 epochs.
We performed a hyperparameter search and fixed the hyperparameters as shown in \cref{tab:geobench_hparams}.
We report the results in terms of their \ac{mAP} for multi-label scene classification, \ac{AA} for multi-class scene classification, \ac{IoU} for semantic segmentation, and F$_1$-score for uni- and cross-modal retrieval.
All scores are reported in \%.
For each \ac{CBIR} task, the task is denoted as \texttt{<Q>$\rightarrow$<R>}, where \texttt{<Q>} denotes the image modality of the query images and \texttt{<R>} denotes the image modality of the retrieved images.
\chadded{For retrieval, the overall feature vector of each image is constructed via \ac{GAP} over the patch representations for \ac{CSMAE}, following \cite{hackstein2024exploring}, and via the \texttt{[CLS]}-token for \ac{CSMoE}, following the ablation in \cref{tab:ablation_token}.}
\begin{table}
\centering
\renewcommand{\arraystretch}{1}
\setlength\tabcolsep{3pt}
\caption{The hyperparameters selected for evaluation on downstream datasets for the comparison of different \acp{FM}.}
\label{tab:geobench_hparams}
\begin{tabular}{llcccr}
\toprule
\multirow{2}{*}{Task} & \multirow{2}{*}{Dataset} & CBIR && Probing & \multirow{2}{*}{Metric}\\
\cmidrule{3-3}
\cmidrule{5-5}
&& $K$ && LR \\
\midrule
\multirow{4}{*}{Classification} & m-bigearthnet  & - && \num{1e-2} & mAP$_\mu$ \\
                     & m-brick-kiln   &  - && \num{3e-3} & \ac{AA} \\
                     & m-so2sat       &  - && \num{1e-2} & \ac{AA} \\
                     & m-eurosat      &  - && \num{5e-2} & \ac{AA} \\
\cmidrule{2-6}
\multirow{2}{*}{Segmentation} & m-cashew-plant &  - && \num{3e-4} & IoU \\
                              & m-SA-crop-type &  - && \num{1e-2} & IoU \\
\cmidrule{2-6}
\multirow{2}{*}{CBIR} & BENv2-14k  & 5 && - & F$_1$ \\
                      & BENv2-243k & 5 && - & F$_1$ \\
\bottomrule
\end{tabular}
\end{table}
\newpagetoggle

\section{Experimental Results}\label{sec:evaluation}
\chadded{
This section presents the experimental results of the proposed \ac{CSMoE} model.
It includes a sensitivity analysis related to its key design choices, a comparison against state-of-the-art \ac{RS} \acp{FM}, and an analysis of the learned expert routing behavior.
}

\subsection{Sensitivity Analysis}

In this subsection, we investigate the impact of \chadded{four} key design choices of the proposed \ac{CSMoE} model \chadded{and its pre-training data}: 1) patch size $\rho$; 2) classification token construction strategy; 3) training duration; \chadded{and 4) our thematic-climatic descriptor-driven sampling strategy, which we compare against random sampling within the same strata.}
\chadded{For each design choice, we report its effect on downstream performance and, where computational cost is affected, \acp{FLOP} to quantify the associated} trade-offs in terms of computational efficiency.

\subsubsection{Patch Size}
\begin{table*}[t]
\centering
\renewcommand{\arraystretch}{0.9}
\setlength\tabcolsep{3pt}
\caption{Comparison of different patch sizes $\rho$ using linear probing \chadded{and their respective pretraining time}. Performance is reported as \ac{mAP} (\%) for m-bigearthnet, \ac{AA} (\%) for m-brick-kiln, m-so2sat, and m-eurosat, and \ac{IoU} (\%) for m-cashew-plant and m-SA-crop-type. \chadded{Pretraining time is reported in GPU-hours, with the hardware used for each run given in parentheses.}}
\label{tab:model_geobench_comparison}
\begin{tabular}{lccccccccrrc}
\toprule
\multirow{2}{*}{Model}&\multirow{2}{*}{\makecell{$\rho$}} & \multicolumn{4}{c}{Classification} && \multicolumn{2}{c}{Segmentation} & \multirow{2}{*}{\makecell{\# Params \aup}} & \multirow{2}{*}{FLOPs \adown} & \multirow{2}{*}{\makecell{\chadded{Pretraining time}\\\chadded{(on hardware)}}\chadded{\adown}}\\
\cmidrule{3-6}
\cmidrule{8-9}
 && m-bigearthnet & m-brick-kiln & m-so2sat & m-eurosat && m-cashew-plant & m-SA-crop-type\\
\midrule
\multirow{4}{*}{CSMoE} & 32 &      62.6  &      93.2  &      44.1  &      84.9  &&      46.0  &      35.8  & \tbf{277M} &  \tbf{2.92G} & \chadded{\tul{56 (A100)}}\\ 
                       & 28 &      65.1  &      93.8  &      46.3  &      84.9  &&      48.3  &      36.7  & \tul{275M} &  \tul{3.67G} & \chadded{\tbf{46 (A100)}}\\ 
                       & 16 & \tbf{66.5} & \tul{94.3} & \tul{48.0} & \tul{86.2} && \tul{55.7} & \tul{38.6} &      271M  &      10.11G  & \chadded{     84 (A100)} \\ 
                       & 14 & \tul{66.0} & \tbf{94.4} & \tbf{49.6} & \tbf{88.3} && \tbf{59.4} & \tbf{39.8} &      271M  &      13.40G  & \chadded{     58 (H200)} \\ 
\bottomrule
\end{tabular}
\end{table*}
To analyze the effect of patch size $\rho$ on the performance and computational cost of the \ac{CSMoE} model, we train four \ac{CSMoE} variants with four different patch sizes $\rho \in \{14, 16, 28, 32\}$.
\Cref{tab:model_geobench_comparison} shows the downstream performance on the scene classification and semantic segmentation datasets, as well as the computational requirements in terms of number of parameters, \acp{FLOP} and \chadded{measured pretraining time} for different patch sizes $\rho$.
 
From the table, one can see that \ac{CSMoE} variants trained with smaller patch sizes ($\rho \in \{14, 16\}$) outperform those using larger patch sizes ($\rho \in \{28, 32\}$) across all classification and segmentation tasks, with particularly large gains on the segmentation datasets m-cashew-plant and m-SA-crop-type. 
For example, \ac{CSMoE} with patch size $\rho = 14$ achieves 59.4\% \ac{IoU} and 39.8\% \ac{IoU}, compared to 46.0\% and 35.8\% for \ac{CSMoE} with patch size $\rho = 32$, on the m-cashew-plant and m-SA-crop-type datasets, respectively. 
A possible explanation is that smaller patches retain finer spatial structures in the embeddings, which may be particularly beneficial for tasks requiring precise delineation of heterogeneous land-cover types.

However, this improvement comes with a substantial increase in inference \acp{FLOP}, from 2.92B at $\rho=32$ to 13.40B at $\rho=14$. \chadded{
We observe that $\rho=16$ offers the best trade-off, retaining most of the downstream performance of $\rho=14$ at notably lower inference \acp{FLOP} (10.11B). 
We additionally report measured pretraining times.
All variants pretrain within 46--84 GPU-hours, which is modest in the context of \acp{FM} in \ac{RS} that often require hundreds~\cite{xiong2024dofa} to thousands~\cite{szwarcman2024prithvi, jakubik2025terramind} of GPU-hours. 
Note that pretraining time does not track \acp{FLOP} directly even at fixed batch size and number of epochs, since wall-clock time also depends on how well the resulting tensor shapes map onto hardware-specific kernel optimizations.
We therefore treat \acp{FLOP} as the hardware-neutral measure and pretraining time as a complementary reference.
The $\rho=14$ variant was trained on different hardware (H200) and its wall-clock time is therefore not directly comparable to the A100 runs and is reported only for completeness.}

\subsubsection{Classification Token Construction Strategy}
To assess how different strategies for constructing the classification token affect the downstream performance, we use the pretrained \ac{CSMoE} model with patch size $\rho = 32$ and train a linear classifier, where we vary the strategy for extracting the input token to the classifier.
As shown in \cref{tab:ablation_token}, directly using the \texttt{[CLS]} token yields the best overall performance across all classification tasks, with 62.6\% \ac{mAP} on m-bigearthnet and 84.9\% \ac{AA} on m-eurosat, outperforming alternatives such as \ac{GAP} over all tokens (60.1\% \ac{mAP}, 82.3\% \ac{AA}) or \ac{GAP} excluding the \texttt{[CLS]} token (60.2\% \ac{mAP}, 82.3\% \ac{AA}). 
It is worth noting that reusing the normalization applied during contrastive pretraining, with or without the projection head, leads to significantly degraded results.
For example, using the normalized (denoted as \enquote{norm. \texttt{[CLS]}}) or the normalized and projected \texttt{[CLS]} token (denoted as \enquote{norm. \& proj. \texttt{[CLS]}}) on m-so2sat yields only 21.9\% \ac{AA}, which is less than half of the un-normalized \texttt{[CLS]} token (denoted as \enquote{only \texttt{[CLS]}}), suggesting a misalignment between the pretraining and probing objectives. 
We would also like to note that the strategy for constructing the classification token does not affect the results on segmentation, as the classification token is not used as a feature for segmentation.
We conclude that while \ac{GAP}-based strategies offer reasonable performance, using the raw \texttt{[CLS]} token remains the most effective and robust choice for classification probing.
However, it is critical for achieving good classification performance that the \texttt{[CLS]} token is used without normalization, although it was trained during pretraining with normalization.
\begin{table}[t]
\centering
\renewcommand{\arraystretch}{0.9}
\setlength\tabcolsep{3pt}
\caption{Comparison of different strategies of constructing the classification token using linear probing on \ac{CSMoE} ($\rho = 32$). Performance is reported as \ac{mAP} (\%) for m-bigearthnet and \ac{AA} (\%) for m-brick-kiln, m-so2sat, and m-eurosat.}
\label{tab:ablation_token}
\begin{tabular}{lcccc}
\toprule
\makecell{Classification Token\\Construction Strategy} & m-bigearthnet & m-brick-kiln & m-so2sat & m-eurosat\\
\midrule
\ac{GAP} w/o. \texttt{[CLS]}   & \tul{60.2} & \tul{89.4} &      40.1  & \tul{82.3} \\
\ac{GAP}                       &      60.1  &      88.9  & \tul{40.2} & \tul{82.3} \\
only \texttt{[CLS]}            & \tbf{62.6} & \tbf{93.2} & \tbf{44.1} & \tbf{84.9} \\
norm. \texttt{[CLS]}           &      40.8  &      82.4  &      21.9  &      69.8  \\
norm. \& proj. \texttt{[CLS]}  &      40.8  &      82.4  &      21.9  &      69.8  \\
\bottomrule
\end{tabular}
\end{table}

\subsubsection{Training Length}
\begin{figure}
    \centering
    \includegraphics[width=0.98\linewidth]{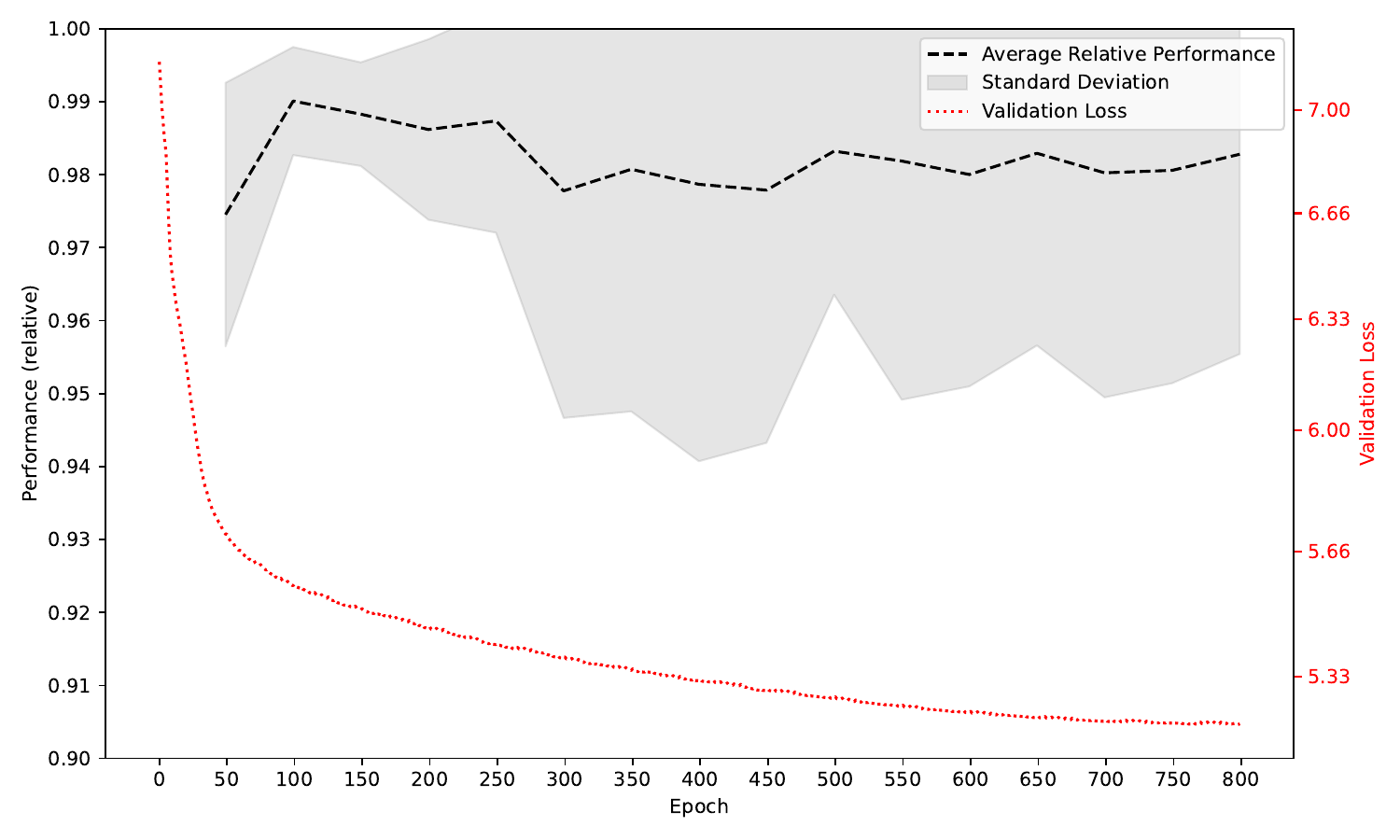}
    \caption{Normalized average performance (\%) and validation loss obtained by \ac{CSMoE} with patch size $\rho = 16$ when pretrained for different numbers of epochs on \ac{mTC} and evaluated via linear and segmentation probing.}
    \label{fig:training_performance}
\end{figure}
To assess the influence of training duration on downstream performance and convergence stability and identify an optimal trade-off between final loss and downstream performance consistency, we evaluate \ac{CSMoE} with patch size $\rho = 16$ trained on \ac{mTC} for up to 800 epochs, where we save one weight checkpoint every 50 epochs.
For each checkpoint, we calculate the normalized performance per dataset as the ratio of the current score to the best score achieved across all epochs for that dataset. 
This normalization allows for averaging across heterogeneous task metrics (e.g., \ac{AA}, \ac{mAP}, and \ac{IoU}).
All checkpoints are evaluated under the same linear probing protocol on the geobench benchmark collection.
\Cref{fig:training_performance} shows the normalized downstream performance as well as the validation loss over the number of epochs trained.
As one can see from the figure, the resulting average normalized performance increases from 97.5\% at epoch 50 to a peak of 99.0\% at epoch 100, while the standard deviation across datasets simultaneously decreases, indicating improved stability. 
After epoch 150 (98.8\% $\pm$ 0.71\%), performance plateaus, while inter-dataset variance gradually increases, with the standard deviation exceeding 3\% from epoch 300 onward. 
The validation loss, however, decreases for the full duration of the training, albeit at a diminishing rate (5.71 at epoch 50, 5.21 at epoch 799). 
We adopt 150 epochs as the default pretraining duration, as it yields near-optimal average performance with low inter-dataset variance, representing the best trade-off between training efficiency and generalization stability.

\subsubsection{\chadded{Sampling Strategy}}
\begin{table*}[t]
\centering
\renewcommand{\arraystretch}{0.9}
\setlength\tabcolsep{3pt}
\caption{\chadded{Comparison of a random stratified subset and an entropy-\chadded{maximized} stratified subset using linear probing and their respective pretraining time. Performance is reported as \ac{mAP} (\%) for m-bigearthnet, \ac{AA} (\%) for m-brick-kiln, m-so2sat, and m-eurosat, and \ac{IoU} (\%) for m-cashew-plant and m-SA-crop-type. \chadded{Pretraining time is reported in GPU-hours, with the hardware used for each run given in parentheses.}}}
\label{tab:sampling_comparison}
\begin{tabular}{lcccccclccc}
\toprule
\multirow{2}{*}{\chadded{Model}}&\multirow{2}{*}{\makecell{\chadded{$\rho$}}} & \multirow{2}{*}{\makecell{\chadded{Sampling}\\\chadded{Algorithm}}} & \multicolumn{4}{c}{\chadded{Classification}} && \multicolumn{2}{c}{\chadded{Segmentation}} & \multirow{2}{*}{\makecell{\chadded{Pretraining time}\\\chadded{(on hardware)}}\chadded{\adown}}\\
\cmidrule{4-7}
\cmidrule{9-10}
 &&& \chadded{m-bigearthnet} & \chadded{m-brick-kiln} & \chadded{m-so2sat} & \chadded{m-eurosat} && \chadded{m-cashew-plant} & \chadded{m-SA-crop-type}\\
\midrule
\multirow{2}{*}{\chadded{CSMoE}} & \multirow{2}{*}{\chadded{32}} & \chadded{random}             & \chadded{     58.1 } & \chadded{     90.8 } & \chadded{\tbf{45.8}} & \chadded{     85.3 } && \chadded{     44.9 } & \chadded{     33.7 } & \chadded{\tbf{2 (H200)}}\\
                                 &                               & \chadded{entropy-maximizing} & \chadded{\tbf{60.3}} & \chadded{\tbf{92.2}} & \chadded{     44.1 } & \chadded{\tbf{85.7}} && \chadded{\tbf{46.4}} & \chadded{\tbf{34.2}} & \chadded{\tbf{2 (H200)}}\\
\bottomrule
\end{tabular}
\end{table*}
\chadded{
To assess the effect of the proposed thematic-climatic descriptor-driven sampling strategy on downstream performance, we compare it against random sampling within the same stratified framework.
To reduce the computational cost of this analysis, we use a scaled-down sampling with $N_s=3$, $T = 50$, and $N_p = 100$.
In this setting, the population size $N_p$ is increased because the small stratum size ($N_s=3$) amplifies the relative effect of mutation in the genetic algorithm, increasing the risk that a stratum collapses to an empty set under the adaptive size constraint. 
A larger population mitigates this risk.
The resulting datasets are used to pretrain \ac{CSMoE} with patch size $\rho=32$ and are evaluated under the same probing protocol described above.
The results of these experiments are given in \cref{tab:sampling_comparison}.
As one can see from the table, the proposed sampling strategy outperforms random sampling on five of six downstream tasks.
For example, on m-bigearthnet, the entropy-maximizing stratified sampling improves 2.2\% \ac{mAP} over the random stratified sampling and on m-cashew-plant, the improvement is 1.5\% \ac{IoU}.
Only on m-so2sat, the random stratified sampling baseline has 1.7\% higher accuracy.
We attribute this to the strong urban bias of m-so2sat, for which the stratification itself may be more influential than the diversity within a stratum. 
We would like to emphasize, however, that both the random sampling and the entropy-maximizing sampling use this stratification.
This results in the random sampling dataset already being highly diverse due to covering the same $\sim$300 combinations of \ac{LULC}-class/\ac{CZ} as the entropy-maximizing sampling.
Still, the consistency of the improvement across heterogeneous downstream tasks spanning land cover, agriculture, and infrastructure suggests that the thematic-climatic descriptor-driven strategy produces a pre-training distribution that transfers more broadly than random sampling within the same strata.
Additionally, we want to highlight that the entropy-maximizing stratified sampling is highly compute efficient and only adds negligible computational cost relative to the pretraining stage it supports.
For example, on the larger dataset ($N_s\approx100$, $T = 2500$, and $N_p = 10$) sampling takes approximately 15 minutes, measured on a machine with 16 CPU cores, while the smaller dataset ($N_s\approx3$, $T = 50$, and $N_p = 100$) takes approximately three minutes to sample on the same machine.
As a limitation, we want to note that due to the randomness of sampling and the filtering of invalid patches, the two datasets slightly differ in size and observed downstream gains therefore may not be attributed exclusively to the sampling strategy.
}

\subsection{Comparison with other Foundation Models}
\begin{figure*}[htbp]
    \centering
    \includegraphics[width=\textwidth]{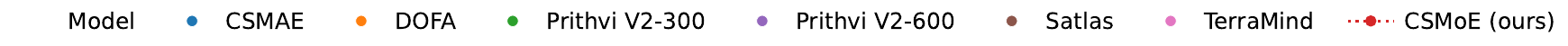}
    \includegraphics[width=0.84\textwidth]{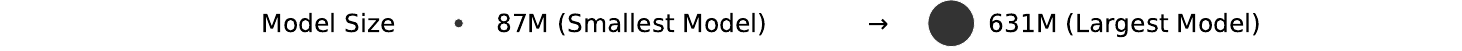}  
    \begin{subfigure}[b]{0.49\textwidth}
        \centering
        \includegraphics[width=\textwidth]{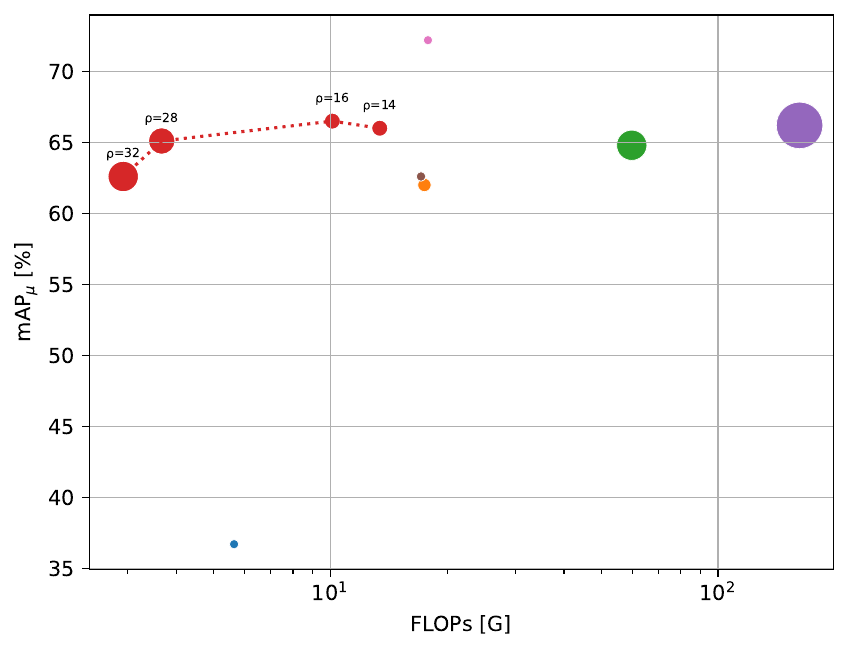}
        \caption{}
        \label{fig:perf_flops_ben}
    \end{subfigure}
    \hfill
    \begin{subfigure}[b]{0.49\textwidth}
        \centering
        \includegraphics[width=\textwidth]{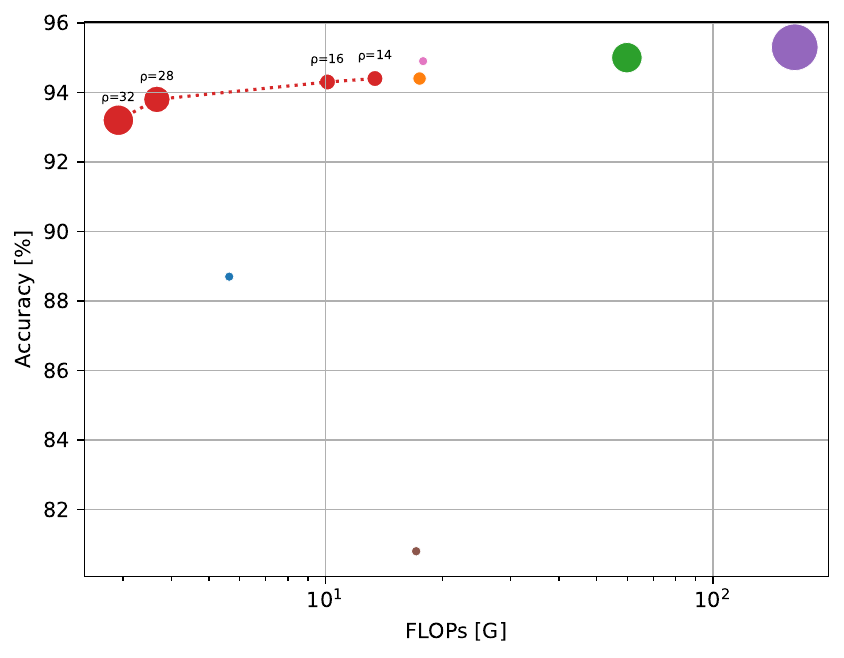}
        \caption{}
        \label{fig:perf_flops_brick}
    \end{subfigure}
    
    \vspace{0.5cm} 
    
    \begin{subfigure}[b]{0.49\textwidth}
        \centering
        \includegraphics[width=\textwidth]{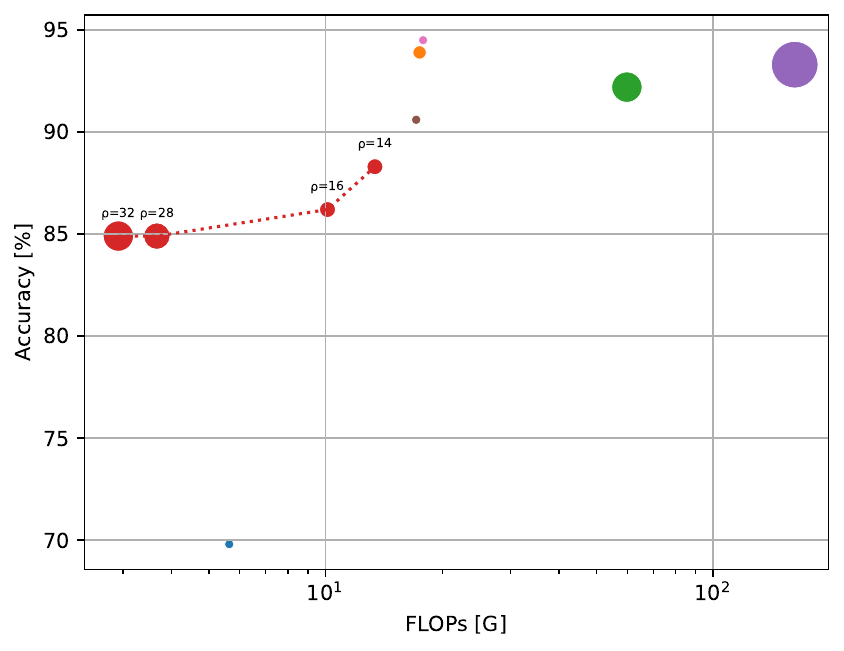}
        \caption{}
        \label{fig:perf_flops_eurosat}
    \end{subfigure}
    \hfill
    \begin{subfigure}[b]{0.49\textwidth}
        \centering
        \includegraphics[width=\textwidth]{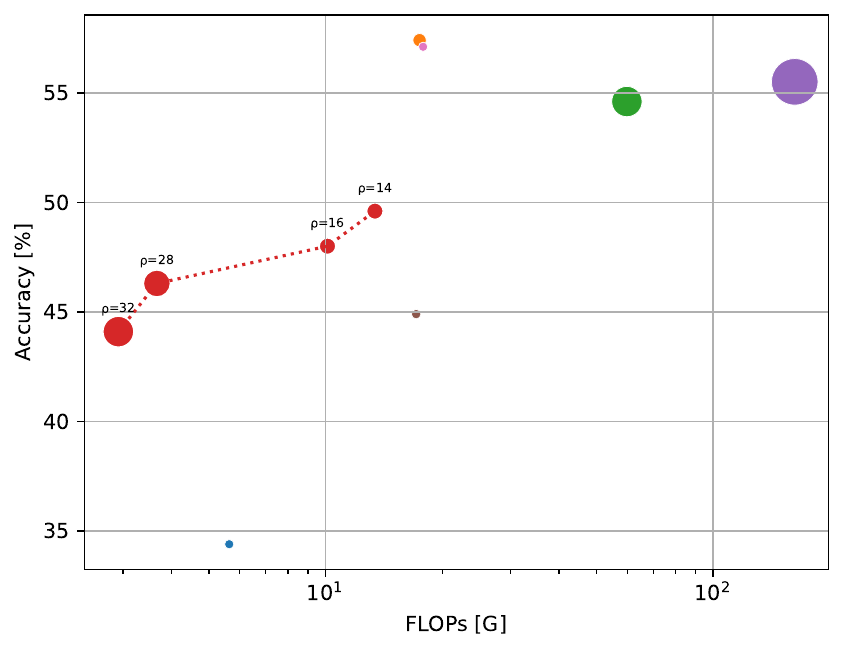}
        \caption{}
        \label{fig:perf_flops_so2sat}
    \end{subfigure}
    \caption{Scene classification results versus the number of \acp{FLOP} on: (a) m-bigearthnet; (b) m-brick-kiln; (c) m-eurosat; and (d) m-so2sat. \chadded{Size of the circle indicates number of parameters in the model.}}
    \label{fig:geobench_cls_over_flops}
\end{figure*}



\begin{figure*}[htbp]
    \centering
    \includegraphics[width=\textwidth]{figures/images/perf_over_flops/models_legend.pdf}
    \includegraphics[width=0.84\textwidth]{figures/images/perf_over_flops/size_legend.pdf}  
    \begin{subfigure}[b]{0.49\textwidth}
        \centering
        \includegraphics[height=6.7cm,width=\textwidth]{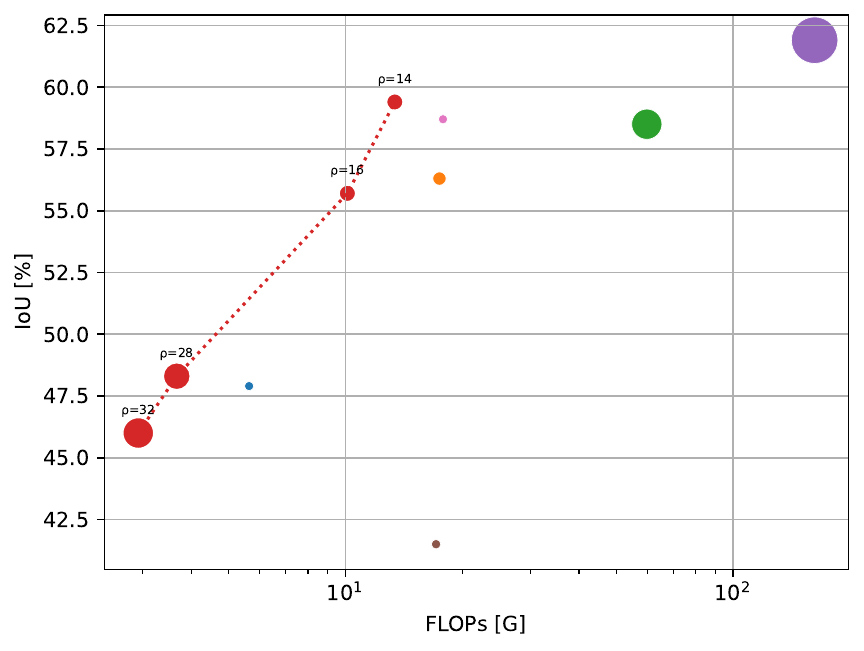}
        \caption{}
        \label{fig:perf_flops_cashew}
    \end{subfigure}
    \hfill
    \begin{subfigure}[b]{0.49\textwidth}
        \centering
        \includegraphics[height=6.7cm,width=\textwidth]{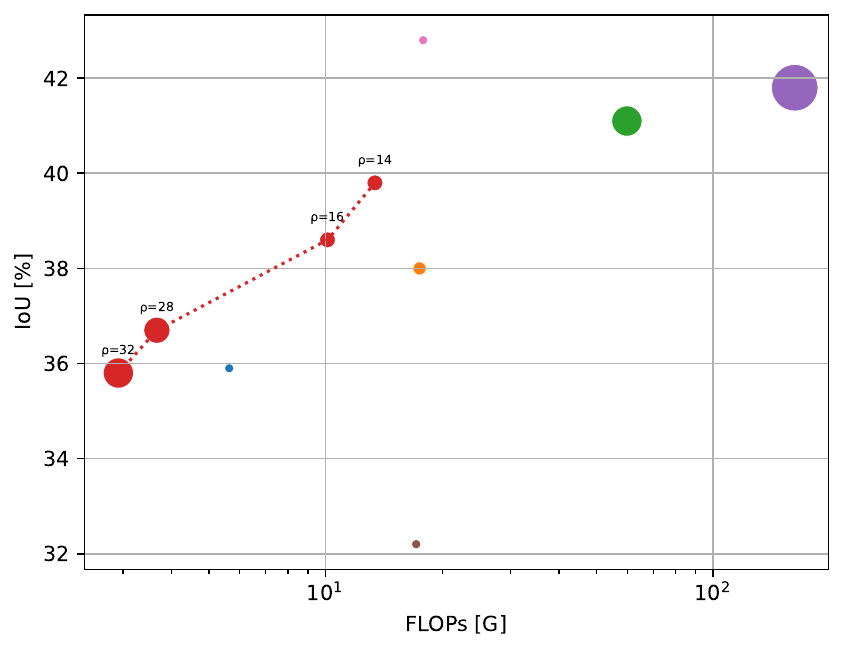}
        \caption{}
        \label{fig:perf_flops_crop}
    \end{subfigure}
    \caption{Semantic segmentation results versus the number of \acp{FLOP} on: (a) m-cashew-plant; and (b) m-SA-crop-type. \chadded{Size of the circle indicates number of parameters in the model.}}
    \label{fig:geobench_seg_over_flops}
\end{figure*}
\begin{figure*}[ht!]
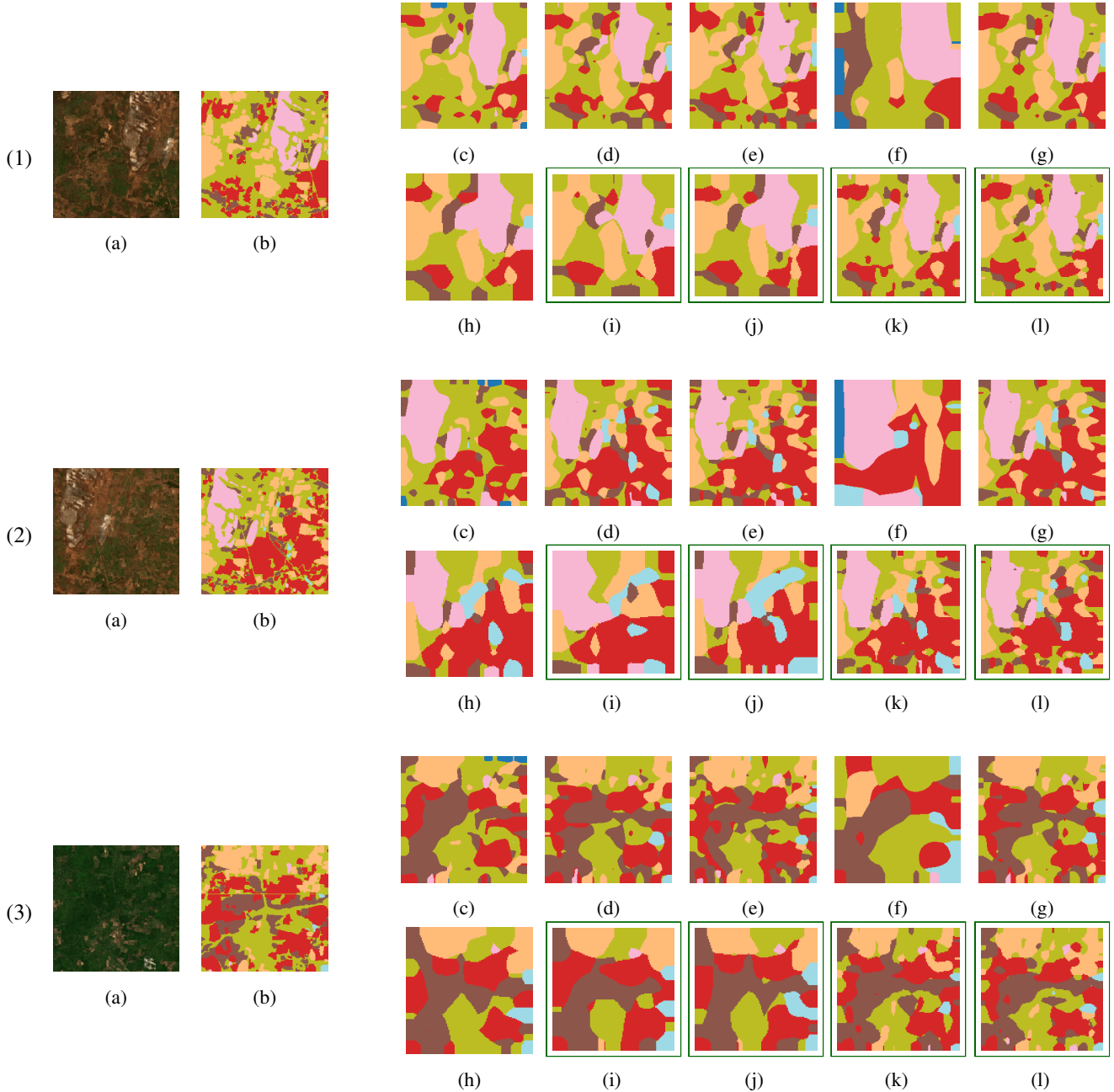

\centering

\newcommand{\imageIDs}{213,242,748}
\newcommand{\modelnames}{dofa,prithvi2_300, prithvi2_600,satlas,SECD_BENv2,ssl4eo,terramind,v014.9,v014.5,v014.7,v014.8}
\newcommand{\getSampleFromID}[1]{%
  \ifthenelse{\equal{#1}{213}}{(1)}{%
  \ifthenelse{\equal{#1}{242}}{(2)}{(3)}}
}
\newcommand{\tightborderimg}[1]{%
\begin{tikzpicture}
  \node[inner sep=0pt, outer sep=0pt, draw=black!60!green, line width=1.5pt] {
    \adjustbox{trim=0px 0px 0px 0px, clip}{\includegraphics[width=\linewidth]{#1}}
  };
\end{tikzpicture}%
}

\foreach \id in \imageIDs {
  \begin{subfigure}[c]{\textwidth}
    \begin{minipage}[c]{0.04\textwidth}
        \centering
        \getSampleFromID{\id}
        \vspace{4mm}
    \end{minipage}%
    \begin{minipage}[c]{0.115\textwidth}
        \centering
        \includegraphics[trim={3 3 3 3},clip,width=\linewidth]{figures/images/segmentation_results/\id/segmentation_\id_input.pdf}
        \caption*{(a)}
    \end{minipage}%
    \hspace{0.01\textwidth}
    \begin{minipage}[c]{0.115\textwidth}
        \centering
        \includegraphics[trim={3 3 3 3},clip,width=\linewidth]{figures/images/segmentation_results/\id/segmentation_\id_label.pdf}
        \caption*{(b)}
    \end{minipage}%
    \hspace{0.02\textwidth}
    \begin{minipage}[c]{0.72\textwidth}
      \centering
    
      \begin{minipage}[t]{0.16\textwidth}
        \includegraphics[trim={3 3 3 3},clip,width=\linewidth]{figures/images/segmentation_results/\id/segmentation_\id_result_dofa.pdf}
        \caption*{(c)}
      \end{minipage}%
      \hspace{2pt}
      \begin{minipage}[t]{0.16\textwidth}
        \includegraphics[trim={3 3 3 3},clip,width=\linewidth]{figures/images/segmentation_results/\id/segmentation_\id_result_prithvi2_300.pdf}
        \caption*{(d)}
      \end{minipage}%
      \hspace{2pt}
      \begin{minipage}[t]{0.16\textwidth}
        \includegraphics[trim={3 3 3 3},clip,width=\linewidth]{figures/images/segmentation_results/\id/segmentation_\id_result_prithvi2_600.pdf}
        \caption*{(e)}
      \end{minipage}%
      \hspace{2pt}
      \begin{minipage}[t]{0.16\textwidth}
        \includegraphics[trim={3 3 3 3},clip,width=\linewidth]{figures/images/segmentation_results/\id/segmentation_\id_result_satlas.pdf}
        \caption*{(f)}
      \end{minipage}%
      \hspace{2pt}
      \begin{minipage}[t]{0.16\textwidth}
        \includegraphics[trim={3 3 3 3},clip,width=\linewidth]{figures/images/segmentation_results/\id/segmentation_\id_result_terramind.pdf}
        \caption*{(g)}
      \end{minipage}
      \newline
      \vspace{1mm}
      \hspace{-8mm}
      \begin{minipage}[t]{0.16\textwidth}
        \includegraphics[trim={3 3 3 3},clip,width=\linewidth]{figures/images/segmentation_results/\id/segmentation_\id_result_SECD_BENv2.pdf}
        \caption*{(h)}
      \end{minipage}%
      \hspace{1pt}
      \begin{minipage}[t]{0.16\textwidth}
        \tightborderimg{figures/images/segmentation_results/\id/segmentation_\id_result_v014.9.pdf}
        \caption*{(i)}
      \end{minipage}%
      \hspace{1pt}
      \begin{minipage}[t]{0.16\textwidth}
        \tightborderimg{figures/images/segmentation_results/\id/segmentation_\id_result_v014.5.pdf}
        \caption*{(j)}
      \end{minipage}%
      \hspace{1pt}
      \begin{minipage}[t]{0.16\textwidth}
        \tightborderimg{figures/images/segmentation_results/\id/segmentation_\id_result_v014.7.pdf}
        \caption*{(k)}
      \end{minipage}%
      \,
      \begin{minipage}[t]{0.16\textwidth}
        \tightborderimg{figures/images/segmentation_results/\id/segmentation_\id_result_v014.8.pdf}
        \caption*{(l)}
      \end{minipage}

    \end{minipage}
  \end{subfigure}

  \vspace{5mm}
}

\caption{Qualitative comparison on three samples from segmentation probing on the m-cashew-plant dataset: (a) input image (RGB); (b) reference maps; and (c) -- (l) obtained segmentation maps from: (c) DOFA; (d) Prithvi V2-300; (e) Prithvi V2-600; (f) Satlas; 
(g) TerraMind; (h) \ac{CSMAE}; and (i-l) \ac{CSMoE} (ours) with patch sizes $\rho=32$, 28, 16 and 14, respectively. 
Our results are highlighted with a green border.}
\label{fig:segmentation_grid}
\end{figure*}

In this subsection, we analyze the effectiveness of the proposed \ac{CSMoE} model for scene classification by comparing it with state-of-the-art \acp{FM} across four scene classification datasets in the geobench benchmark collection. 
\cref{fig:geobench_cls_over_flops} shows the corresponding scene classification results in terms of \ac{mAP} for m-bigearthnet and \ac{AA} for the remaining datasets with respect to the required number of \acp{FLOP}. 
One can see that the \ac{CSMoE} model variants consistently achieve high classification performance across all datasets while requiring significantly fewer \acp{FLOP} compared to most existing approaches. 
As an example, on the m-bigearthnet dataset, all \ac{CSMoE} model variants reach an \ac{mAP} score comparable to Prithvi~V2-600 while requiring an order of magnitude fewer \acp{FLOP} (see \cref{fig:perf_flops_ben}). 
On m-eurosat (\cref{fig:perf_flops_eurosat}) and m-brick-kiln (\cref{fig:perf_flops_brick}), the \ac{CSMoE} model variants yield an \ac{AA} close to or even surpassing \acp{FM} such as the \ac{CSMAE} and Satlas models, while maintaining a much lower computational budget. 
In particular, on the m-brick-kiln, all of our model variants achieve an \ac{AA} comparable to much larger models such as Prithvi V2-300 and TerraMind, despite operating with considerably fewer \acp{FLOP}. 
On the more challenging m-so2sat dataset (\cref{fig:perf_flops_so2sat}), even the most lightweight \ac{CSMoE} model variant that uses a patch size of $\rho = 32$ outperforms the baseline \ac{CSMAE} model, whereas the \ac{CSMoE} model variants with smaller patch sizes narrow the gap to the best-performing \acp{FM} while maintaining high computational efficiency.
We would like to note that the \acp{FLOP} reported in the figures are shown in logarithmic scale to better visualize the trade-off between computational complexity and performance across models. 
As a result, differences in \acp{FLOP} may appear visually less pronounced. 
In general, our model variants with a smaller patch size ($\rho \in \{16, 14\}$) tend to achieve better performance across datasets, except on the m-bigearthnet dataset, where the variant using a patch size of $\rho = 16$ yields a slightly higher \ac{mAP} than the variant using a patch size of  $\rho = 14$.

We observe similar trends in semantic segmentation, as shown in \cref{fig:geobench_seg_over_flops}, which presents results on m-cashew-plant (\cref{fig:perf_flops_cashew}) and m-SA-crop-type (\cref{fig:perf_flops_crop}). 
All of our \ac{CSMoE} model variants demonstrate strong performance while operating at significantly reduced computational cost. 
For instance, on m-cashew-plant, the \ac{CSMoE} model variant with patch size $\rho = 14$ achieves an \ac{IoU} of 59.4\%, outperforming TerraMind (58.8\%) and Prithvi~V2-300 (58.5\%) and closely approaching the performance of Prithvi~V2-600 (61.9\%), while requiring substantially fewer \acp{FLOP} (13.40G vs. 17.84G, 59.85G, and 162.18G, respectively). 
On the m-SA-crop-type dataset, the \ac{CSMoE} model variant with patch size $\rho = 14$ similarly outperforms or closely matches the performance of mid-sized \acp{FM} like \ac{CSMAE}, Satlas, and DOFA. 
In line with our results on scene classification, the \ac{CSMoE} model variants using smaller patch sizes generally yield stronger results, with the variant that uses a patch size of $\rho \in \{14, 16\}$ consistently outperforming their counterparts with larger patch sizes. 
These trends are further illustrated in \cref{fig:segmentation_grid}, which compares model outputs across three scenes from the m-cashew-plant dataset. 
From top to bottom, each example shows the true color representation of the input image (a), the reference map (b), predictions from other \acp{FM} (c–h), followed by our \ac{CSMoE} model variants with patch sizes $\rho=32$, $\rho=28$, $\rho=16$ and $\rho=14$ (i–l).
The results reveal that especially our model variants with smaller patch sizes (k, l) produce cleaner and more coherent segmentation maps, often recovering fine structures and class boundaries more accurately than the less efficient baseline \acp{FM}.

\begin{table*}[!htb]
\centering
\renewcommand{\arraystretch}{0.9}
\setlength\tabcolsep{3pt}
\caption{F$_1$-scores (\%) obtained by \ac{CSMAE} and \ac{CSMoE} with different patch sizes $\rho$ on uni-modal and cross-modal \ac{CBIR} when the image sets of BENv2-14k and BENv2-243k are considered.}
\label{tab:retrieval_results}
\begin{tabular}{lc ccccc c ccccc cc}
\toprule
\multirow{5}{*}{Model} & \multirow{5}{*}{$\rho$} & \multicolumn{11}{c}{Training Set} & \multirow{5}{*}{\# Params \aup} & \multirow{5}{*}{FLOPs \adown}\\
\cmidrule{3-13}
&& \multicolumn{5}{c}{BENv2-14k} && \multicolumn{5}{c}{BENv2-243k} \\
\cmidrule{3-7}
\cmidrule{9-13}
&& \multicolumn{2}{c}{Uni-Modal CBIR} && \multicolumn{2}{c}{Cross-Modal CBIR} && \multicolumn{2}{c}{Uni-Modal CBIR} && \multicolumn{2}{c}{Cross-Modal CBIR}\\
\cmidrule{3-4}
\cmidrule{6-7}
\cmidrule{9-10}
\cmidrule{12-13}
&& \soso & \stst && \sost & \stso && \soso & \stst && \sost & \stso \\
\midrule
CSMAE \cite{hackstein2024exploring} 
             & 16 &      60.71 &       68.62  &&      36.20  &      42.31  &&      56.71  &      63.85  &&      29.79  &      26.33  &       87M  &       5.64G  \\ 
\cmidrule{2-15}
\multirow{4}{*}{CSMoE (ours)} 
             & 32 &      63.59  &      71.04  &&      35.16  &      45.01  &&      60.02  &      68.17  && \tul{31.39} & \tbf{31.86} & \tbf{277M} &  \tbf{2.92G} \\ 
             & 28 &      63.92  &      72.24  && \tul{39.91} &      39.50  &&      60.88  &      69.23  && \tbf{32.02} &      25.51  & \tul{275M} &  \tul{3.67G} \\ 
             & 16 & \tul{66.58} & \tul{72.33} &&      35.93  & \tbf{50.74} && \tul{62.77} & \tul{70.66} &&      25.83  & \tul{30.15} &      271M  &      10.11G  \\ 
             & 14 & \tbf{67.32} & \tbf{73.34} && \tbf{42.22} & \tul{46.76} && \tbf{63.77} & \tbf{71.38} &&      29.54  &      28.35  &      271M  &      13.40G  \\ 
\bottomrule
\end{tabular}
\end{table*}

\chadded{
For \ac{CBIR}, we evaluate the \ac{CSMAE} and the proposed \ac{CSMoE} model on the BENv2-14k and BENv2-243k sets of images from BigEarthNet-v2.
\Cref{tab:retrieval_results} shows the corresponding F$_1$-scores, the required number of model parameters and the \acp{FLOP} when the two training sets are considered for uni-modal and cross-modal \ac{CBIR}.
By assessing the table, one can observe that in the uni-modal \ac{CBIR} scenario all of our \ac{CSMoE} model variants surpass the \ac{CBIR} performance of \ac{CSMAE} on both retrieval tasks and on both sets of images.
As an example, on BENv2-14k, the \ac{CSMoE} model variant using a patch size of $\rho=14$ achieves 67.32\% F$_1$-score on the \soso task and 73.34\% on the \stst task, which is 6.61\% and 4.72\% higher than \ac{CSMAE}.
On the BENv2-243k set of images, the same \ac{CSMoE} model variant again achieves the highest F$_1$-scores of the evaluated models on the \soso and \stst tasks with 63.77\% and 71.38\%, compared to 56.71\% and 63.85\% for \ac{CSMAE}.
In greater detail, one can also observe that our most lightweight model variant, which uses a patch size of $\rho=32$, surpasses \ac{CSMAE} on all uni-modal tasks while operating with 48\% fewer \acp{FLOP}.
}

\chadded{
In the more challenging cross-modal \ac{CBIR} scenario, all models achieve lower F$_1$-scores.
In this scenario, the \ac{CSMoE} model variants do not surpass \ac{CSMAE} on both retrieval tasks at the same time.
However, one can observe from the table that a \ac{CSMoE} model variant that achieves a lower F$_1$-score than \ac{CSMAE} on one retrieval task always achieves a higher F$_1$-score on the other.
As an example, on BENv2-14k, the \ac{CSMoE} model variant using a patch size of $\rho=32$ achieves 35.16\% on the \sost task compared to 36.20\% for \ac{CSMAE}, while achieving 45.01\% on the \stso task compared to 42.31\%.
The variant using a patch size of $\rho=28$ shows the opposite behavior, achieving 39.91\% on the \sost task compared to 36.20\%, but 39.50\% on the \stso task compared to 42.31\%.
The same behavior is observed on BENv2-243k, where the variant using a patch size of $\rho=16$ achieves 25.83\% on the \sost task compared to 29.79\% for \ac{CSMAE}, while achieving 30.15\% on the \stso task compared to 26.33\%.
This is due to the fact that the two cross-modal retrieval tasks are complementary views of the same shared embedding space.
Accordingly, a model variant for which this space is aligned more favorably for one retrieval task is aligned less favorably for the other.
The \ac{CSMoE} model variant using a patch size of $\rho=14$ is the exception, as it surpasses \ac{CSMAE} on both retrieval tasks on BENv2-14k with 42.22\% and 46.76\%, compared to 36.20\% and 42.31\%.
On BENv2-243k, the highest F$_1$-scores are achieved by our model variants using patch sizes of $\rho=28$ on the \sost task with 32.02\% and $\rho=32$ on the \stso task with 31.86\%, which surpass \ac{CSMAE} while operating with 35\% and 48\% fewer \acp{FLOP}, respectively.
We would also like to note that \ac{CSMAE} was explicitly optimized for \ac{CBIR}, whereas \ac{CSMoE} was pretrained for broad downstream transfer without task-specific optimization.
}

\chadded{
For both sets of images, smaller patch sizes lead to higher uni-modal \ac{CBIR} performance, in line with our results on scene classification and semantic segmentation.
On the cross-modal retrieval tasks, this behavior is not observed, and the patch size achieving the highest F$_1$-score instead varies with the retrieval task and the considered set of images.
This is due to the larger domain gap between the two modalities, which is addressed by the representation capacity of the encoders rather than by the spatial granularity associated with a specific patch size.
In particular, \ac{CSMoE} with patch size $\rho=32$, which is our model variant that requires the lowest number of \acp{FLOP}, remains competitive on all tasks and achieves the highest F$_1$-score of all evaluated models on the \stso task on BENv2-243k.
These results show the effectiveness of the proposed \ac{CSMoE} model for \ac{CBIR} under restricted computational constraints.
}

\subsection{\chadded{MoE Learning Behavior}}\label{sec:expert_analysis}
\chadded{
To understand how \ac{CSMoE}'s experts contribute to the observed performance, we analyze the internal routing behavior of the expert layers along three axes: 
i)~the per-expert usage variance across images; 
ii)~the correlation between per-expert usage and land-cover labels, controlled for low-level intensity statistics via partial correlation and assessed against a label-permutation null; and 
iii)~the diversity of the learned expert weights, measured via pairwise cosine similarity. 
For i) and ii), we consider 512 random but fixed images per modality from BigEarthNet-v2.0~\cite{clasen2025refinedbigearthnet}. 
To further test whether sharper routing induces modality specialization, we additionally evaluate downstream retrieval performance for four \ac{CSMoE} variants that differ only in the dispatch temperature used during pretraining ($0.1$, $0.25$, $0.5$, and $1.0$).
All variants use $\rho=32$.
}
\begin{table}[htb]
    \centering
    \renewcommand{\arraystretch}{0.9}
    \setlength\tabcolsep{3pt}
    \caption{\chadded{Expert usage analysis for the \ac{CSMoE} model on BigEarthNet-v2.0~\cite{clasen2025refinedbigearthnet}. Usage variance is the per-expert usage variance across images, with low values indicating near-uniform routing. $r_{\text{label}}$ is the maximum absolute usage--label correlation; $r_{\text{stat}}$ the maximum absolute usage--intensity-statistic correlation; $r_{\text{partial}}$ the usage--label correlation after controlling for per-band intensity statistics; and null the 95th percentile of the permutation null for $r_{\text{partial}}$. The high $r_{\text{stat}}$ and the collapse of $r_{\text{partial}}$ show that the apparent usage--label correlation is largely explained by input magnitude.}}
    \label{tab:expert_usage}
    \begin{tabular}{llcccccc}
    \toprule
    \makecell{\chadded{Input}\\\chadded{Modality}} & \chadded{Encoder} & \chadded{Layer} & \chadded{usage var.} & \chadded{$r_{\text{label}}$} & \chadded{$r_{\text{stat}}$} & \chadded{$r_{\text{partial}}$} & \chadded{null} \\
    \midrule
    \multirow{6}{*}{\chadded{\ac{S2}}}
      & \multirow{4}{*}{\chadded{$e_{MS}^{S2}$}} & \chadded{1} & \chadded{1.08e$^{-6}$} & \chadded{0.422} & \chadded{0.855} & \chadded{0.192} & \chadded{0.194} \\
                                               & & \chadded{2} & \chadded{3.14e$^{-7}$} & \chadded{0.469} & \chadded{0.780} & \chadded{0.233} & \chadded{0.182} \\
                                               & & \chadded{3} & \chadded{1.07e$^{-7}$} & \chadded{0.292} & \chadded{0.856} & \chadded{0.186} & \chadded{0.151} \\
                                               & & \chadded{4} & \chadded{6.44e$^{-8}$} & \chadded{0.386} & \chadded{0.665} & \chadded{0.178} & \chadded{0.171} \\\cmidrule{2-8}
      & \multirow{2}{*}{\chadded{$e_{CS}$}}      & \chadded{1} & \chadded{1.80e$^{-8}$} & \chadded{0.263} & \chadded{0.602} & \chadded{0.112} & \chadded{0.164} \\
                                               & & \chadded{2} & \chadded{7.42e$^{-9}$} & \chadded{0.429} & \chadded{0.651} & \chadded{0.133} & \chadded{0.164} \\
    \midrule
    \multirow{6}{*}{\chadded{\ac{S1}}}
      & \multirow{4}{*}{\chadded{$e_{MS}^{S1}$}} & \chadded{1} & \chadded{1.69e$^{-7}$} & \chadded{0.640} & \chadded{0.938} & \chadded{0.352} & \chadded{0.148} \\
                                               & & \chadded{2} & \chadded{3.71e$^{-7}$} & \chadded{0.515} & \chadded{0.740} & \chadded{0.386} & \chadded{0.134} \\
                                               & & \chadded{3} & \chadded{6.17e$^{-7}$} & \chadded{0.467} & \chadded{0.503} & \chadded{0.403} & \chadded{0.157} \\
                                               & & \chadded{4} & \chadded{1.87e$^{-8}$} & \chadded{0.477} & \chadded{0.616} & \chadded{0.305} & \chadded{0.146} \\\cmidrule{2-8}
      & \multirow{2}{*}{\chadded{$e_{CS}$}}      & \chadded{1} & \chadded{1.83e$^{-8}$} & \chadded{0.427} & \chadded{0.559} & \chadded{0.205} & \chadded{0.143} \\
                                               & & \chadded{2} & \chadded{5.20e$^{-9}$} & \chadded{0.548} & \chadded{0.843} & \chadded{0.258} & \chadded{0.131} \\
    \bottomrule
    \end{tabular}
\end{table}

\chadded{
In the first analysis, we assess the per-expert usage variance across images. 
It is very low across all layers (on the order of $10^{-6}-10^{-9}$ as can be seen in \cref{tab:expert_usage}), confirming that routing is close to uniform and that the experts act as a soft ensemble rather than as selectively activated specialists.
In the second analysis, we assess the correlation between per-expert usage and land-cover labels.
The maximum usage-label correlation reaches up to $r_{\text{label}}=0.64$.
However, \cref{tab:expert_usage} shows that expert usage is far more strongly correlated with low-level intensity statistics ($r_{\text{stat}}$ up to $0.938$) than with class labels.
After controlling for per-band intensity statistics, the partial correlation for \ac{S2} falls to or near the null in most layers, whereas a consistent residual remains for \ac{S1} ($|r|$ up to $\approx0.40$ compared to a null of $\approx0.15$). 
We attribute this residual to the physical relationship between \ac{SAR} backscatter intensity and land cover rather than to semantic expert specialization.
}

\chadded{
In the third analysis, we assess the diversity of the learned expert weights. 
\Cref{tab:expert_cosine_sim} shows the pairwise cosine similarity between expert weights within each of the ten \ac{MoE} layers.
As one can see from the table, the experts remain near-orthogonal, with pairwise cosine similarities within $0 \pm 0.007$ for the modality-specific experts and within $0 \pm 0.035$ for the shared experts.
This confirms that the experts learn distinct functions rather than collapsing into redundant copies.
The higher variance observed for the shared experts is likely due to them processing more varied inputs originating from different modalities, requiring them to differentiate more strongly.
}

\begin{table}[htb]
    \centering
    \renewcommand{\arraystretch}{0.9}
    \setlength\tabcolsep{3pt}
    \caption{\chadded{Pairwise cosine similarities between expert weights with respect to model depth. The similarities are close to zero, indicating that the expert functions remain distinct. The shared experts (in encoder $e_{CS}$) exhibit a higher variance, possibly because they process more varied inputs (i.e., different data modalities) and therefore have to differentiate more.}}
    \label{tab:expert_cosine_sim}
    \begin{tabular}{lcccc}
    \toprule
    \multirow{2}{*}{\chadded{Encoder}} & \multirow{2}{*}{\chadded{Layer}} & \multicolumn{3}{c}{\chadded{Cosine Similarity}}\\
    \cmidrule{3-5}
    & & {\chadded{min.}} & {\chadded{mean}} & {\chadded{max.}} \\
    \midrule
      \multirow{4}{*}{\chadded{$e_{MS}^{S1}$}} & \chadded{1} & \chadded{-0.004} & \chadded{ 0.000} & \chadded{ 0.004} \\
                                               & \chadded{2} & \chadded{-0.004} & \chadded{-0.001} & \chadded{ 0.003} \\
                                               & \chadded{3} & \chadded{-0.005} & \chadded{ 0.000} & \chadded{ 0.005} \\
                                               & \chadded{4} & \chadded{-0.002} & \chadded{ 0.001} & \chadded{ 0.005} \\\midrule
      \multirow{4}{*}{\chadded{$e_{MS}^{S2}$}} & \chadded{1} & \chadded{-0.002} & \chadded{-0.000} & \chadded{ 0.004} \\
                                               & \chadded{2} & \chadded{-0.003} & \chadded{ 0.000} & \chadded{ 0.002} \\
                                               & \chadded{3} & \chadded{-0.004} & \chadded{ 0.001} & \chadded{ 0.005} \\
                                               & \chadded{4} & \chadded{-0.007} & \chadded{-0.000} & \chadded{ 0.005} \\\midrule
      \multirow{2}{*}{\chadded{$e_{CS}$}} & \chadded{1} & \chadded{-0.016} & \chadded{ 0.002} & \chadded{ 0.019} \\
                                          & \chadded{2} & \chadded{-0.032} & \chadded{ 0.004} & \chadded{ 0.034} \\
    \bottomrule
    \end{tabular}
\end{table}

\begin{table}[htb]
    \centering
    \renewcommand{\arraystretch}{0.9}
    \setlength\tabcolsep{3pt}
    \caption{\chadded{F$_1$-scores (\%) obtained by \ac{CSMoE} with different dispatch temperatures on uni-modal and cross-modal \ac{CBIR} when the image set of BENv2-14k is considered. All variants use $\rho=32$.}}
    \label{tab:temperature_ablation}
    \begin{tabular}{lccccc}
    \toprule
    \multirow{2}{*}{\makecell{\chadded{Dispatch}\\\chadded{Temperature}}} & \multicolumn{2}{c}{\chadded{Uni-Modal CBIR}} && \multicolumn{2}{c}{\chadded{Cross-Modal CBIR}} \\
    \cmidrule{2-3}
    \cmidrule{5-6}
    & \chadded{\soso} & \chadded{\stst} && \chadded{\sost} & \chadded{\stso} \\
    \midrule
    \chadded{1.0}  & \chadded{     61.15 } & \chadded{     70.12 } && \chadded{\tbf{37.24}} & \chadded{     41.99 } \\
    \chadded{0.5}  & \chadded{\tbf{63.59}} & \chadded{\tbf{71.04}} && \chadded{     35.16 } & \chadded{\tbf{45.01}} \\
    \chadded{0.25} & \chadded{     58.44 } & \chadded{     67.55 } && \chadded{     37.14 } & \chadded{     44.60 } \\
    \chadded{0.1}  & \chadded{     60.48 } & \chadded{     70.64 } && \chadded{     30.37 } & \chadded{     38.70 } \\
    \bottomrule
    \end{tabular}
\end{table}

\chadded{
To further assess whether sharper dispatch is beneficial, we evaluate the effect of the pretraining dispatch temperature on downstream retrieval performance.
\Cref{tab:temperature_ablation} shows the corresponding F$_1$-scores on BigEarthNet-v2.0 for the uni-modal and cross-modal retrieval tasks.
As one can see from the table, lowering or increasing the dispatch temperature during pretraining does not improve retrieval performance.
The variant using a dispatch temperature of $0.5$, which is the one used in our final model, achieves the best F$_1$-score across three of the four retrieval tasks. 
It reaches 63.59\% and 71.04\% on the uni-modal \soso and \stst tasks and 35.16\% and 45.01\% on the cross-modal \sost and \stso tasks, respectively.
Both lower- and higher-temperature variants yield no consistent improvement and, in most cases, substantially degrade performance.
These observations indicate that sharper routing during pretraining reduces performance while providing no specialization benefit.
}

\chadded{
In summary, our analysis shows that \ac{CSMoE} behaves as a soft ensemble of diverse experts combined with near-uniform routing weights, rather than as a redundant or semantically specialized mixture.
We therefore conclude that the performance gains of the proposed \ac{CSMoE} model arise from its architectural integration and soft-mixing capacity, and not from a discrete specialization of individual experts.
}
\newpagetoggle

\section{Conclusion}\label{sec:conclusion}
In this paper, for the first time in \ac{RS}, we have investigated the effectiveness of injecting Soft \acp{MoE} into \acp{FM} to decrease their computational requirements while retaining their representational capacity. 
\chadded{
To this end, we have improved computational efficiency along two axes: i)~model efficiency; and ii)~training-data efficiency.
Model efficiency is achieved by a general adaptation for efficient multi-modal processing, which injects a Soft \ac{MoE} into an \ac{RS} \ac{FM} through the integration of expert routing, encoder architecture modifications, and the regularization of training objectives.
We have applied this adaptation to \ac{CSMAE}, resulting in the first compute-efficient multi-modal \ac{FM} in \ac{RS}, which we call \ac{CSMoE}.
Training-data efficiency is achieved by a thematic-climatic descriptor-driven sampling strategy, which leverages climate zones and thematic products to ensure thematic-climatic diversity while promoting spatial diversity through genetic optimization.
We have trained our \ac{CSMoE} model using a training set constructed with this strategy.
}
To evaluate our \ac{CSMoE} model, we have carried out extensive experiments on scene classification, semantic segmentation, and \ac{CBIR}, while comparing the \ac{CSMoE} model with other state-of-the-art \acp{FM}. 
Experimental results demonstrate the effectiveness of \ac{CSMoE}, achieving performance comparable or superior to existing \acp{FM} while requiring significantly fewer \acp{FLOP}. 

In our experiments, we have also investigated the effects of patch size, strategies for constructing the classification token, the number of pretraining epochs, and the sampling strategy used to construct the pretraining set on the required computational resources and the resulting downstream performance of our \ac{CSMoE} model.
From this sensitivity analysis, we have derived guidelines to select the appropriate \ac{CSMoE} model variant under different computational requirements in \ac{RS} as follows:
\begin{itemize}
    \item If computational resources are severely constrained, the \ac{CSMoE} model variant with patch size $\rho=32$ can be selected due to its \chadded{low computational requirements} while maintaining competitive performance.
    \item For the best trade-off between efficiency and performance, the \ac{CSMoE} model variant with patch size $\rho=16$ can be selected for its high performance,  especially in segmentation tasks.
    \item For the highest downstream performance, the \ac{CSMoE} model variant that is using a patch size of $\rho=14$ can be selected, achieving the highest performance of the \ac{CSMoE} model variants while maintaining a lower computational complexity than the existing \acp{FM}.
\end{itemize}
\chadded{We have also analyzed the routing behavior of the experts in our \ac{CSMoE} model.
The results show that the experts operate as a soft ensemble with near-uniform routing weights and near-orthogonal parameters, rather than as semantically specialized units.
Accordingly, the performance of the proposed \ac{CSMoE} model is achieved by the integration of the Soft \ac{MoE} mechanism itself, without requiring the individual experts to specialize.}

We would like to note that, in this paper, we realized our Soft \ac{MoE} adaptation in the context of one specific \ac{RS} \ac{FM}, namely \ac{CSMAE}. 
However, the proposed adaptation can be applied to most existing \acp{FM} in \ac{RS} to improve their representational capacity and reduce their computational complexity. 
Additionally, the proposed adaptation is not limited to models using masked image modeling as the learning objective, but can also be used for models using contrastive learning or any other self-supervised learning objective. 
As future works, we plan to: 
i)~extend our model to be able to process other modalities such as video and text; and 
ii)~increase the efficiency of the proposed model by including task-specific expert pruning or expert creation at runtime.
\newpagetoggle

\printbibliography[title=References]



\end{document}